\documentclass[10pt,twocolumn,letterpaper]{article}
\usepackage[pagenumbers]{cvpr} %

\usepackage{amsmath,amssymb,booktabs,multirow,array}
\usepackage{capt-of}
\usepackage{tikz}
\usetikzlibrary{arrows.meta,positioning,fit,backgrounds,decorations.pathreplacing,calc,shapes.multipart}
\definecolor{shaS}{HTML}{0072B2}
\definecolor{shaA}{HTML}{009E73}
\definecolor{cvprblue}{rgb}{0.21,0.49,0.74}
\definecolor{draftred}{HTML}{A43632}
\usepackage[breaklinks,colorlinks,allcolors=cvprblue]{hyperref}
\usepackage{float}

\newcommand{\mat}[1]{\mathbf{#1}}

\title{Localizing and Fixing the Instability of Hadamard Rotations\\on 4-bit Vision Transformers}
\hypersetup{pdftitle={Localizing and Fixing the Instability of Hadamard Rotations on 4-bit Vision Transformers}}
\author{%
Keshan Yang$^{1}$\quad Xiaowen Duan$^{2}$\quad Zelin Yang$^{1}$\quad Lijian Zhu$^{3}$\\
Zhonghua Peng$^{4}$\quad Hongliang Luo$^{1}$\quad Xing Wei$^{1}$\thanks{Corresponding author.}\\[4pt]
\small $^{1}$Xi'an Jiaotong University\\
\small $^{2}$The Chinese University of Hong Kong, Shenzhen\\
\small $^{3}$Shenzhen Heils Zhongcheng Technology Co., Ltd.\\
\small $^{4}$Shenzhen University\\[2pt]
\small \texttt{yangkeshan@stu.xjtu.edu.cn}, \texttt{weixing@xjtu.edu.cn}
}
\hypersetup{%
pdfauthor={Keshan Yang, Xiaowen Duan, Zelin Yang, Lijian Zhu, Zhonghua Peng, Hongliang Luo, Xing Wei},%
pdfsubject={Preprint}%
}

\begin{document}
\maketitle

\begin{abstract}
Hadamard rotations make 4-bit weights and activations (W4A4) practical for large language models, whose activations tend to concentrate in a few outlier channels. However, on some vision transformers, such as DINOv3, the unsigned Hadamard that these methods apply online loses much of the accuracy. We trace this loss to the GELU output entering the second MLP layer (fc2): its channel means do not cancel, so a full-width Hadamard collects each token's channel sum into one coordinate that sets the quantization range. The same coordinate appears in GELU and ReLU language models and is absent in SwiGLU models. Random signs remove it but cost a separate pass at fc2. We instead propose a grouped Hadamard (GH) that rotates the fc2 input in groups of 64 channels. GH bounds the coordinate and runs inside the kernel of the first MLP layer (fc1) that produces the input, at 0.7 to 2.3\% of encoder latency. On ImageNet, inserted before the unchanged quantizers of four post-training quantization (PTQ) methods, GH improves all 28 method-model pairs. On the SAM2.1 image encoders, our INT4 engine built on GH and two frozen-weight corrections loses at most 0.4 mIoU against floating point and runs 1.12 to 1.27 times faster than the W8A8 path of the same engine, and 1.78 to 2.78 times faster than TensorRT FP16. Code and the INT4 engine are available at \url{https://github.com/kohillyang/SGHAQuant}.
\end{abstract}

\section{Introduction}
\begin{figure}[t]
  \centering
  \includegraphics[width=\columnwidth]{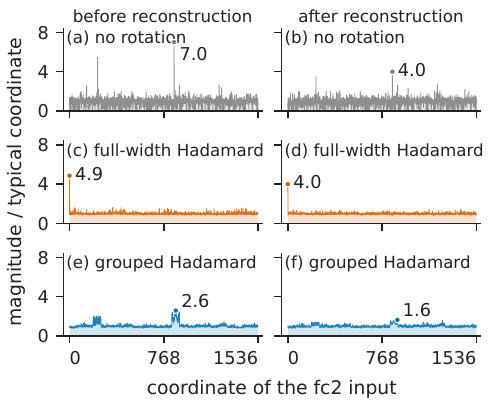}
  \caption{\textbf{Outlier channels and the collected channel sum at the fc2 input} (DeiT-S, block~5). Each bar is the 99.9th-percentile magnitude of one coordinate over tokens, relative to the median coordinate. Left column: before reconstruction. Right column: after the weight reconstruction of FIMA-Q~\cite{fimaq} with that transform (Table~\ref{tab:plugin} checkpoints). Both columns on 32 of FIMA-Q's calibration images. (a,~b)~Without rotation, a few outlier channels set the range, 7.0 and 4.0 times the median. (c,~d)~A full-width Hadamard rotation collects the GELU channel sum of each token into its first coordinate, 4.9 and 4.0 times the median. (e,~f)~With 64-channel groups, every coordinate stays within 2.6 and 1.6 times the median.}
  \label{fig:meanconc}
\end{figure}
\begin{table}[b]
\footnotesize
\setlength{\tabcolsep}{1.5pt}
\caption{\textbf{Mechanism ladder at W4A4}. FH: the full-width Hadamard~\cite{quarot,spinquant}, GH the grouped Hadamard. Group width: at qkv and fc1 / at fc2, $K$ the full width, -- none. Random signs: $\pm1$ per channel before FH at the three inputs~\cite{quipsharp}. Static centering: a fixed per-channel mean from calibration data subtracted before FH at fc2. Per-token centering: each token's mean over channels subtracted before FH at fc2. DeiT-S, ViT-S: ImageNet top-1 (\%) under RepQ-ViT. Hiera-T: SAM2.1 point-prompt mIoU in simulation with our engine's quantizers. EVA-02-S: SwiGLU, top-1 under RepQ-ViT, only qkv, fc1 and fc2 quantized.}
\label{tab:ladder}
\begin{tabular*}{\columnwidth}{@{}l@{\extracolsep{\fill}}ccccc@{}}
\toprule
Transform & group width & DeiT-S & ViT-S & Hiera-T & EVA-02-S\\
\midrule
floating point & & 79.85 & 81.39 & 60.68 & 85.72\\
none & & 68.93 & 65.24 & 50.79 & 71.08\\
\addlinespace[2pt]
FH at qkv, fc1 & $K$/-- & 71.56 & 69.12 & 54.01 & 75.76\\
GH at qkv, fc1 & 128/-- & 71.53 & 68.27 & 53.53 & 75.90\\
\addlinespace[2pt]
GH at both & 128/64 & 74.57 & 72.61 & 56.27 & 82.66\\
FH at both & $K$/$K$ & 70.07 & 60.54 & 47.97 & 82.65\\
FH + random signs & $K$/$K$ & 74.97 & 72.99 & 57.31 & 83.56\\
FH + static centering & $K$/$K$ & 73.50 & 26.65 & 57.27 & 82.88\\
FH + per-token centering & $K$/$K$ & 75.16 & 73.11 & 58.02 & 82.81\\
\bottomrule
\end{tabular*}
\end{table}

Vision transformers~\cite{vit} are the image encoders of foundation models such as SAM and SAM2.1~\cite{sam,sam2}, and in dense prediction the encoder dominates the inference cost~\cite{sparsesam}. Post-training quantization (PTQ) reduces this cost using a small calibration set without retraining. With 4-bit weights and activations (W4A4), the linear layers run on INT4 tensor cores, which the widely deployed Ampere and Ada GPUs, including consumer cards, provide at twice the INT8 throughput. A 4-bit quantizer, however, has only 16 levels and is sensitive to a few channels, known as outlier channels, whose activations are much larger than the rest and set the quantizer's range for a whole token~\cite{llmint8,outliersupp}.

Language models typically handle outlier channels with equivalent transforms, which change the operands of a layer but not its output: a per-channel scaling moves part of the activation range into the weights~\cite{smoothquant}, and an orthogonal rotation mixes the channels so that no single channel sets the range~\cite{quarot,spinquant}. At the input of the second MLP layer (fc2), QuaRot and SpinQuant apply the rotation online as an unsigned Hadamard transform, the fast Walsh--Hadamard transform at power-of-two widths. On vision transformers, however, rotations turn out to be unstable: they help some models and lose much accuracy on others. ERQ reruns SpinQuant at W4A4 on seven ViT, DeiT and Swin models and finds that it loses at least 19 points of top-1~\cite{erqtpami}. Park and Lee measure fixed Hadamard rotations, with and without random signs, at 15.9 to 16.3\% on DeiT-S under per-tensor scales~\cite{parklee2026rotate}, and RegCache reports that a naive QuaRot damages vision encoders and that its adapted QuaRot still loses 12.5 points on CLIP at W6A6~\cite{regcache}. In our runs with the quantizers of RepQ-ViT~\cite{repqvit}, a full-width Hadamard rotation also raises DeiT-S by 1.1 points and lowers ViT-S by 4.7 (Table~\ref{tab:ladder}). On DINOv3-L~\cite{dinov3} it drops top-1 to chance level (Sec.~\ref{sec:dinov3}).

In this paper, we trace the instability to the large nonzero channel means of the GELU output~\cite{gelu} that enters fc2: the all-ones row of the unsigned Hadamard sums them, so each token's channel sum lands in one coordinate that sets the quantization range. The same coordinate also appears at the GELU- and ReLU-fed fc2 inputs of language models such as GPT-2 and OPT, and is absent at the SwiGLU inputs of EVA-02 and Qwen2.5 (Sec.~\ref{sec:where}).

To fix the instability, we split the channels into consecutive groups of a fixed width and apply a Hadamard rotation within each group, which we call the grouped Hadamard (GH). Each group collects only its own partial sum, which stays small, so the collected coordinate is bounded. More importantly, a group of 64 channels fits inside the epilogue of the GEMM of the first MLP layer (fc1), which produces the fc2 input, so the fc2 rotation runs at a small fraction of the encoder latency (Table~\ref{tab:cost}). We further find that the groups that hold an outlier channel remain larger than the rest after grouping. The weight reconstruction of FIMA-Q~\cite{fimaq} and LS-ViT~\cite{lsvit} can lower them (Fig.~\ref{fig:meanconc}e,~f; Sec.~\ref{sec:host}), which offsets the weaker outlier suppression of the narrower width. GH is therefore compatible with existing reconstruction methods.

Our contributions are as follows.
\begin{itemize}[leftmargin=*,itemsep=1pt,topsep=2pt]
  \item We localize the instability of the unsigned Hadamard rotation on W4A4 vision transformers to the GELU-fed fc2 input, where the all-ones row collects each token's channel sum into one coordinate that sets the quantization range.
  \item We propose GH, a grouped Hadamard rotation that rotates each group of channels independently, bounds this coordinate and is fused into the kernel that produces the input. Inserted before the unchanged quantizers of four PTQ methods, GH improves all 28 method-model pairs on ImageNet (Sec.~\ref{sec:plugin}); the gain also carries over to the SAM ViT and DINOv3 encoders (Secs.~\ref{sec:samv1} and~\ref{sec:dinov3}).
  \item We measure end-to-end encoder latency on real GPUs. Built on GH, the W4A4 path of our engine keeps the SAM2.1 Hiera~\cite{hiera} and DINOv3-L encoders near floating-point accuracy and runs them 1.12 to 1.30$\times$ faster than its W8A8 path (Secs.~\ref{sec:sam21} and~\ref{sec:dinov3}).
\end{itemize}

\section{Related Work}
\label{sec:related}
\paragraph{Rotations in vision models.}
Rotation-based PTQ was developed for language models: QuIP and QuIP\# rotate the weights with random orthogonal or randomized Hadamard factors~\cite{quip,quipsharp}, and subsequent methods rotate the activations as well. The activation transforms differ in two costs, arithmetic and memory traffic. In arithmetic, a Sylvester Hadamard of width $n$ costs $O(n\log n)$ per token through the fast Walsh--Hadamard transform (QuaRot, SpinQuant, CentriQ and the blocks of BRQ and ConvRot~\cite{quarot,spinquant,centriq,brq,convrot}), a dense block-diagonal transform $O(nb)$ for blocks of width $b$ (DuQuant~\cite{duquant}), a Kronecker-structured one $O(n^{3/2})$ (FlatQuant~\cite{flatquant}), and a diagonal scaling folds into the weights (SmoothQuant, RepQ-ViT~\cite{smoothquant,repqvit}). The larger cost is memory traffic. On vision transformers a transform that runs as a pass of its own over the activation costs a sizable share of the encoder latency and can cost more than the W4A4 speedup over W8A8 (Sec.~\ref{sec:cost}). This splits the transforms into two classes. The full-row Hadamards of QuaRot, SpinQuant, HadaNorm~\cite{hadanorm} and CentriQ, with or without random signs or mean centering, a dense random rotation, FlatQuant's Kronecker transform and DuQuant's zigzag permutation across all blocks need the pass; FlatQuant's fused kernel and CentriQ's single-pass kernel are that pass. Diagonal scalings of the SmoothQuant family fold into the weights, but alone they are usually not enough on vision transformers (Sec.~\ref{supp:sq} of the supplement). Rotations on vision transformers have been tried in some existing work. Park and Lee report that a fixed Hadamard falls to 16\% on DeiT-S at W4A4 with per-tensor scales and that adding RepQ-ViT's reparameterization on top of their rotation lowers accuracy against the rotation alone~\cite{parklee2026rotate}. In this paper, inspired by ConvRot, by analyzing the collected coordinate on vision transformers we locate it at one input and show that GH, as a low-cost variant of the Hadamard rotation, improves accuracy on a wide range of vision transformers.

\paragraph{Post-training quantization of vision transformers.}
PTQ methods for vision transformers differ in how much they fit and at what cost. The cheapest need a few calibration passes. Most of them fit the quantizer to the activation distribution without optimizing the weights: twin uniform grids~\cite{ptq4vit}, a log quantizer for softmax outputs~\cite{fqvit}, extended to GELU outputs~\cite{adalog}, per-channel scales reparameterized into the LayerNorm (LN) weights~\cite{repqvit} or instance-aware channel groups~\cite{igqvit}. GPTQ instead updates the weights layer by layer in closed form~\cite{gptq}, and QwT and QwT-v2 add a closed-form output compensation to several hosts~\cite{qwt,qwtv2}. Block reconstruction optimizes the weight rounding against floating-point block outputs for thousands to tens of thousands of iterations per block~\cite{brecq,qdrop,isvit}, with refined Hessian approximations~\cite{aphqvit,fimaq,lsvit}, and Joint PTQ fits per-channel scales and shifts on the post-LN inputs, the quantizer parameters and a weight refinement end to end on 10k real images, in about one hour for ViT-S~\cite{jointptq}. SAM-specific PTQ sits in this last tier. It handles the bimodal post-key distribution~\cite{ptq4sam}, adds hybrid log-uniform quantizers and channel grouping~\cite{ahcptq}, extends them with a weight regularization and a log quantizer for attention scores~\cite{ahcqsam}, or reconstructs the cross-attention branches and evaluates with ground-truth box prompts~\cite{carsam}. On large models, however, this reconstruction can be very time-consuming; for example, in our runs AHCQ-SAM takes 39 h on SAM ViT-L and 58 h on ViT-H on one RTX 4090. In this paper, we reach competitive accuracy on the SAM encoders at a cost from the cheapest tier: the grouped rotation alone, with round-to-nearest weights, keeps the SAM ViT-B, ViT-L and ViT-H encoders within 1.1 AP of floating point at the full coverage of PTQ4SAM (Sec.~\ref{sec:samv1}), and with two frozen-weight corrections fitted in minutes the SAM2.1 encoders stay within 0.4 mIoU (Sec.~\ref{sec:sam21}).

\paragraph{Low-bit execution.}
TensorRT on Ampere and Ada GPUs, QAttn, ViDiT-Q and QServe provide W8A8 or W4A8 execution but no W4A4~\cite{qattn,viditq,qserve}. With 4-bit activations, Atom and SVDQuant provide W4A4 kernels for language and diffusion models~\cite{atom,svdquant}, QuantVGGT reports W4A4 about 1.15$\times$ faster than W8A8 on the VGGT aggregator on an RTX 4090, with the DINOv2 tokenizer and the heads in floating point in its released code~\cite{quantvggt}, and SparseSAM states that SAM W4A4 rarely gives a measurable GPU speedup~\cite{sparsesam}. No existing work reports a W4A4 speedup over W8A8 measured on a GPU for the SAM and DINOv3 encoders.

\section{Why Are Unsigned Rotations Unstable on Vision Transformers}
\label{sec:where}
\begin{figure*}[t]
  \centering
  \resizebox{\textwidth}{!}{\begin{tikzpicture}[
  font=\footnotesize,
  >={Stealth[length=1.6mm]},
  blk/.style={draw, rounded corners=2pt, minimum height=0.95cm, minimum width=1.5cm, align=center, inner sep=2.5pt},
  fixedb/.style={blk, draw=black!45, fill=black!4},
  sca/.style={blk, draw=shaS, fill=shaS!9, line width=0.8pt},
  aff/.style={blk, draw=shaA, fill=shaA!10, line width=0.8pt},
  note/.style={align=center, font=\scriptsize, text=black!75},
  lab/.style={font=\footnotesize\bfseries, anchor=west, inner sep=0pt},
  plus/.style={circle, draw=black!60, inner sep=0pt, minimum size=3.2mm}
]
\node[lab] at (-0.35,1.85) {(a) One quantized layer};
\node (x) at (0,0) {$\mathbf{X}$};
\node[sca, right=0.9cm of x] (S) {\textbf{S}cale\\$\mathbf{D}=\mathrm{diag}(d)$};
\node[fixedb, right=0.9cm of S] (H) {\textbf{G}rouped\\\textbf{H}adamard};
\node[fixedb, right=0.9cm of H] (Q) {4-bit\\quantizer $Q_a$};
\node[fixedb, right=0.9cm of Q, minimum width=2.35cm] (G) {INT4 GEMM\\$Q_w(\mathbf{H}^{\top}\mathbf{D}^{-1}\mathbf{W})$};
\node[aff, right=0.9cm of G] (A) {\textbf{A}ffine\\gain $g$, bias $\beta$};
\node[right=0.9cm of A] (y) {$\widehat{\mathbf{Y}}$};
\foreach \a/\b in {x/S, S/H, H/Q, Q/G, G/A, A/y} \draw[->] (\a) -- (\b);
\begin{scope}[on background layer]
  \node[draw=black!50, densely dashed, rounded corners=4pt, fit=(S)(A), inner xsep=7pt, inner ysep=6pt] (unit) {};
\end{scope}
\draw[decorate, decoration={brace, amplitude=4pt}, shaS, line width=0.6pt]
  ([yshift=9pt]S.north west) -- ([yshift=9pt]S.north east) node[midway, above=5pt, note, text=shaS] {input side: before the quantizers};
\draw[decorate, decoration={brace, amplitude=4pt}, shaA, line width=0.6pt]
  ([yshift=9pt]A.north west) -- ([yshift=9pt]A.north east) node[midway, above=5pt, note, text=shaA] {output side: after the GEMM};

\node[lab] at (-0.35,-1.45) {(b) S, GH and A in a transformer block: a fused implementation};
\begin{scope}[font=\scriptsize,
  kern/.style={rectangle split, rectangle split horizontal, rectangle split parts=2, draw=black!60, line width=0.6pt,
               minimum height=0.8cm, inner xsep=3pt, align=center, rectangle split part fill={black!4,white}},
  kern3/.style={kern, rectangle split parts=3, rectangle split part fill={black!4,white,white}},
  fpk/.style={draw=black!35, densely dashed, minimum height=0.8cm, inner xsep=3pt, align=center, text=black!65},
  ub/.style={decorate, decoration={brace, amplitude=4pt, mirror}, line width=0.6pt, draw=black!60},
  ul/.style={font=\scriptsize, align=center, text=black!80}]
\coordinate (in) at (x.east |- 0,-2.60);
\node[kern, anchor=west] (k1) at ([xshift=0.6cm]in) {LN$_{\color{shaS}\mathbf{S}}$\nodepart{two}\textbf{GH}};
\node[kern, right=0.29cm of k1] (k2) {GEMM\nodepart{two}{\color{shaA}$\mathbf{A}$}};
\node[fpk, right=0.29cm of k2] (k3) {attention};
\node[fpk, right=0.29cm of k3] (k4) {proj};
\node[plus, right=0.29cm of k4] (p1) {$+$};
\node[kern, right=0.29cm of p1] (k5) {LN$_{\color{shaS}\mathbf{S}}$\nodepart{two}\textbf{GH}};
\node[kern3, right=0.29cm of k5] (k6) {GEMM\nodepart{two}{\color{shaA}$\mathbf{A}$}, GELU\nodepart{three}{\color{shaS}$\mathbf{S}$}, \textbf{GH}};
\node[kern, right=0.29cm of k6] (k7) {GEMM\nodepart{two}{\color{shaA}$\mathbf{A}$}};
\node[plus, right=0.29cm of k7] (p2) {$+$};
\coordinate (out) at (y.west |- p2);
\foreach \a/\b in {in/k1, k1/k2, k2/k3, k3/k4, k4/p1, p1/k5, k5/k6, k6/k7, k7/p2, p2/out} \draw[->] (\a) -- (\b);
\draw[->] ($(in)!0.5!(k1.west)$) -- ++(0,0.75) -| (p1.north);
\draw[->] ($(p1.east)!0.5!(k5.west)$) -- ++(0,0.75) -| (p2.north);
\draw[ub] ([yshift=-3pt]k1.south west) -- ([yshift=-3pt]k2.south east) node[midway, below=6pt, ul] {quantized layer: qkv};
\draw[ub] ([yshift=-3pt]k5.south west) -- ([yshift=-3pt,xshift=-1pt]k6.two split south) node[midway, below=6pt, ul] {quantized layer: fc1};
\draw[ub] ([yshift=-3pt,xshift=1pt]k6.two split south) -- ([yshift=-3pt]k7.south east) node[midway, below=6pt, ul] {quantized layer: fc2};
\node[note, text=shaS, text width=3.4cm, anchor=north] at ($(k1.south)!0.5!(k2.south)+(0,-0.95)$) {$\mathbf{S}$ of qkv and fc1: folded into the LayerNorm weight and bias};
\node[note, text=shaA, text width=3.6cm, anchor=north] at ($(k3.south)!0.5!(k4.south)+(0,-0.95)$) {$\mathbf{A}$: folded into the GEMM scale and bias};
\node[note, text=shaS, text width=3.6cm, anchor=north] at ($(k6.three south)+(0,-0.95)$) {$\mathbf{S}$ of fc2: applied in the fc1 epilogue};
\end{scope}
\end{tikzpicture}}
  \caption{\textbf{S, GH and A in a Transformer block.} (a)~One quantized layer: a channel \textbf{S}caling and a \textbf{G}rouped \textbf{H}adamard rotation before the host quantizer, and a per-channel \textbf{A}ffine map (gain and bias) on the INT4 GEMM output. (b)~The kernels of one block; braces mark the kernels of each quantized layer, and a box ``X\,$|$\,Y'' is one kernel that performs X and then Y on chip. Each GH stage ends with the 4-bit quantizer $Q_a$ of (a). Attention and the output projection keep the precision of the host.}
  \label{fig:method}
\end{figure*}
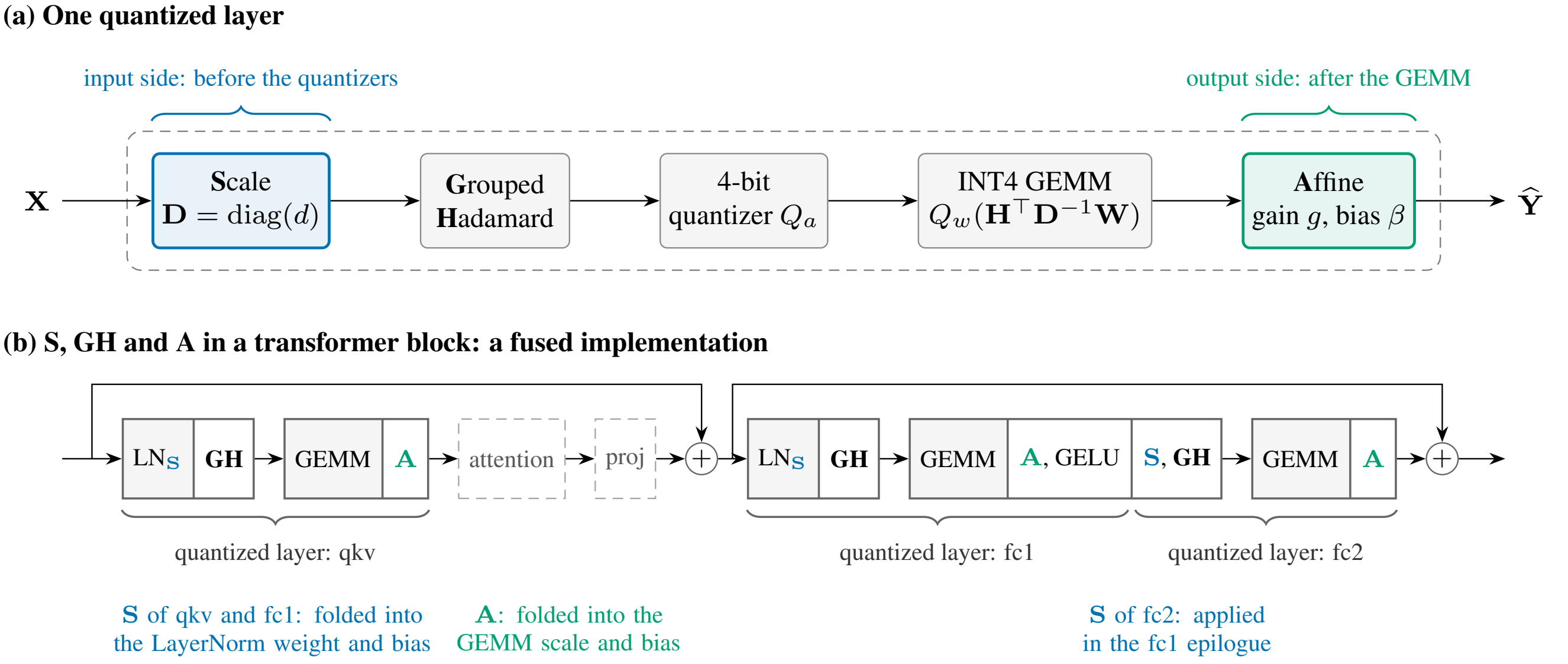

\begin{table}[t]
\footnotesize
\setlength{\tabcolsep}{3.2pt}
\caption{\textbf{The collected coordinate on four models} (floating point, median over blocks). $K$: width of the input. Share: the collected coordinate relative to the largest coordinate of the rotated input (1.00: it sets a max-scaled range). Bits: levels of a 4-bit grid spent above the 99.9th percentile.}
\label{tab:diag}
\begin{tabular*}{\columnwidth}{@{}ll@{\extracolsep{\fill}}r@{\extracolsep{\fill}}c@{\extracolsep{\fill}}cc@{\extracolsep{\fill}}cc@{\extracolsep{\fill}}cc@{}}
\toprule
& & & none & \multicolumn{2}{c}{grouped H} & \multicolumn{2}{c}{full-width H} & \multicolumn{2}{c}{random}\\
\cmidrule(lr){4-4}\cmidrule(lr){5-6}\cmidrule(lr){7-8}\cmidrule(lr){9-10}
Model & input & $K$ & bits & share & bits & share & bits & share & bits\\
\midrule
DeiT-S & qkv & 384 & 1.25 & 0.60 & 0.55 & 0.18 & 0.58 & 0.76 & 0.52\\
       & fc2 & 1536 & 1.80 & 0.84 & 0.92 & \textbf{1.00} & \textbf{2.01} & 0.64 & 0.89\\
ViT-S  & qkv & 384 & 1.01 & 0.65 & 0.55 & 0.20 & 0.53 & 0.74 & 0.59\\
       & fc2 & 1536 & 1.75 & 0.96 & 0.83 & \textbf{1.00} & \textbf{1.80} & 0.70 & 0.85\\
ViT-B  & qkv & 768 & 1.32 & 0.79 & 0.58 & 0.26 & 0.62 & 0.74 & 0.64\\
       & fc2 & 3072 & 2.22 & 0.91 & 0.89 & \textbf{1.00} & \textbf{2.21} & 0.64 & 0.91\\
DeiT-T & qkv & 192 & 0.64 & 0.60 & 0.53 & 0.12 & 0.54 & 0.74 & 0.53\\
       & fc2 & 768 & 1.30 & 0.83 & 0.83 & \textbf{1.00} & \textbf{1.07} & 0.70 & 0.83\\
\bottomrule
\end{tabular*}
\end{table}

This section discusses why the unsigned rotation is unstable, before Sec.~\ref{sec:method} states the transform. Table~\ref{tab:ladder} reads top to bottom as a causal chain. Its none row is each host's W4A4 quantizers without a transform: RepQ-ViT, with its LayerNorm reparameterization, on DeiT-S, ViT-S and EVA-02-S, and the quantizers of our engine on Hiera-T. Against no rotation, a full-width Hadamard gains 1.1 points on DeiT-S and loses 4.7 on ViT-S. The collected sum changes from token to token. Subtracting each token's own mean before the rotation recovers the loss on DeiT-S, ViT-S and Hiera-T, whereas the static calibration mean leaves ViT-S at 26.65 (Table~\ref{tab:ladder}). Every transform that breaks or bounds the all-ones row recovers the loss, but GH is the only one of them that needs no separate pass over the fc2 input (Sec.~\ref{sec:rule}).

\paragraph{Localization.}
The inputs of the query-key-value projection (qkv) and of fc1 are LN outputs, and there the full width is harmless. Applied at these two inputs only, a full-width rotation is as accurate as a grouped one on all four models, and better on ViT-S (Table~\ref{tab:ladder}). Table~\ref{tab:diag} shows why. LN centers each token over its channels, so at qkv and fc1 the all-ones row no longer collects the token mean. What it collects comes from the LN bias and gain alone, and is small on these models. The fc2 input, however, is a GELU output, bounded below, and its channel means do not cancel, so the channel sum is large. At full width the collected coordinate sets the maximum on DeiT-S, ViT-S, ViT-B and DeiT-T and costs up to 2.2 of the 4 bits. Groups of 64 channels or a random rotation cost less than one. The same pattern holds on Swin, on the four Hiera encoders and at the GELU- and ReLU-fed fc2 inputs of GPT-2, Pythia, OPT and CLIP ViT-B/16. At the SwiGLU inputs of EVA-02 and Qwen2.5 the channel means cancel and the coordinate does not appear, and the full-width and the grouped rotation reach the same accuracy on EVA-02-S (Table~\ref{tab:ladder}) and on EVA-02-B (Sec.~\ref{supp:diag} of the supplement). The coordinate thus follows the activation function.

\section{Method}
\label{sec:method}
In this section, we introduce our method in detail. Sec.~\ref{sec:transforms} writes the quantized layer with the three transforms: a channel scaling S, a grouped Hadamard rotation GH and a per-channel affine map A. Sec.~\ref{sec:group} starts from the plain Hadamard rotation and shows how we group it into GH. On hosts without weight reconstruction, GH alone leaves part of the error, so Sec.~\ref{sec:host} designs S before the rotation and A after the GEMM as two lightweight corrections, which together with GH form SGHAQuant. Finally, Sec.~\ref{sec:rule} analyzes the overhead of the three transforms.

\subsection{The Quantized Layer}
\label{sec:transforms}
Consider a linear layer $\mat{Y}=\mat{X}\mat{W}+\mathbf{1}_M b$, where $\mat{X}\in\mathbb{R}^{M\times K}$ holds $M$ tokens of $K$ channels, $\mat{W}\in\mathbb{R}^{K\times N}$ has $N$ output channels, $b\in\mathbb{R}^{1\times N}$ is the bias and $\mathbf{1}_M\in\mathbb{R}^{M\times 1}$ is the all-ones vector. A PTQ method quantizes $\mat{X}$ and $\mat{W}$ with quantizers $Q_a$ and $Q_w$. We insert a positive channel scaling $\mat{D}=\mathrm{diag}(d)$ with $d\in\mathbb{R}^{K}$ (S) and a block-diagonal matrix $\mat{H}\in\mathbb{R}^{K\times K}$ that rotates each group of $n$ channels by a normalized Hadamard matrix (GH) before the quantizers, and a per-output-channel gain $g\in\mathbb{R}^{N}$ and bias $\beta\in\mathbb{R}^{1\times N}$ (A) after the product (Fig.~\ref{fig:method}a):
\begin{equation}
  \widehat{\mat{Y}}=\big(Q_a(\mat{X}\mat{D}\mat{H})\,Q_w(\mat{H}^{\top}\mat{D}^{-1}\mat{W})\big)\,\big(\mat{I}+\mathrm{diag}(g)\big)+\mathbf{1}_M(b+\beta).
  \label{eq:module}
\end{equation}
Without quantization the two factors multiply back to $\mat{X}\mat{W}$, because $\mat{H}$ is orthogonal. On hosts with block reconstruction we use GH alone, i.e., $\mat{D}=\mat{I}$ and $g=\beta=0$.

\subsection{Grouped Rotation}
\label{sec:group}
A full-width Hadamard rotation~\cite{quarot,spinquant} multiplies the $K$ channels of each token by the normalized Hadamard matrix $\mat{H}_K$ of Sylvester's construction~\cite{quipsharp}, $\mat{H}_1=(1)$ and $\mat{H}_{2m}=2^{-1/2}\bigl(\begin{smallmatrix}\mat{H}_m&\mat{H}_m\\\mat{H}_m&-\mat{H}_m\end{smallmatrix}\bigr)$. Its all-ones first row collects the channel sum of Sec.~\ref{sec:where}, and the fc1 epilogue, which produces the fc2 input, holds only one output tile of a token, a small fraction of the fc2 row (Fig.~\ref{fig:method}b). We therefore split the $K$ channels into independent groups of $n$ and rotate each group by $\mat{H}_n$:
\begin{equation}
  \mat{H}=\begin{pmatrix}\mat{H}_n&&\\&\ddots&\\&&\mat{H}_n\end{pmatrix}.
  \label{eq:gh}
\end{equation}
Each of the $K/n$ groups collects only the sum of its own $n$ channels, divided by $\sqrt{n}$. The group width is a power of two that fits the tile; we use a narrower group at fc2 than at qkv and fc1, whose LN-fed inputs are insensitive to the collected coordinate (Sec.~\ref{sec:where}). Sec.~\ref{sec:setup} gives the widths.

\subsection{A Lightweight Correction for Hosts without Reconstruction}
\label{sec:host}
\begin{figure}[t]
  \centering
  \includegraphics[width=\columnwidth]{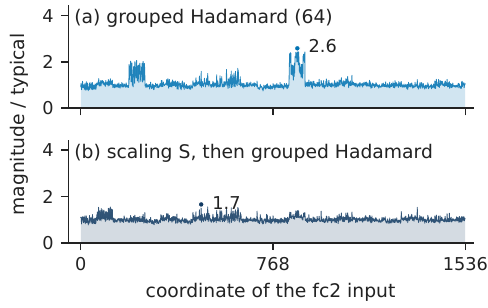}
  \caption{\textbf{The channel scaling S before the grouped rotation} (DeiT-S, block~5, fc2 input; statistic as in Fig.~\ref{fig:meanconc}). (a)~Under GH, the group that holds an outlier channel rises to 2.6 times the median. (b)~With S before the rotation, the largest coordinate is 1.7 times the median.}
  \label{fig:s}
\end{figure}

GH bounds the collected coordinate but not the outlier channels themselves: under GH, the group that holds an outlier channel still stands out (Fig.~\ref{fig:s}a). Hosts with block reconstruction lower this group when they refit the weights (Fig.~\ref{fig:meanconc}e and f). A calibration-only host has no such step, and existing block reconstruction can be time-consuming on models the size of SAM2.1 (Sec.~\ref{sec:related}). For these hosts we add two lightweight corrections, inspired by SmoothQuant~\cite{smoothquant} and QwT-v2~\cite{qwtv2}: a closed-form channel scaling S before the rotation and a per-channel affine map A after the GEMM. Both are fitted once on the calibration images, with the pretrained weights frozen, and fold into the weights and biases.

S lowers this group in closed form before the rotation (Fig.~\ref{fig:s}b). It rescales each channel by its standard deviation $\sigma_c$ over the calibration tokens,
\begin{equation}
  d_c\propto\sigma_c^{-1/2},
  \label{eq:closed}
\end{equation}
normalized by a constant so that the product of the $d_c$ over the channels is one, then clipped to $\log d_c\in[-1,1]$. The exponent $1/2$ is the $\alpha=0.5$ of SmoothQuant~\cite{smoothquant}, applied as a closed-form rule as in QuIP~\cite{quip}.

A corrects the output side. It keeps the per-output-channel gain $g$ and bias $\beta$ of QwT-v2~\cite{qwtv2}, but instead of a per-layer closed form it fits them jointly per block, with the pretrained weights frozen, by minimizing
\begin{equation}
 \mathcal{L}_\ell=\mathbb{E}_{\mat{X}_\ell}\,
 \frac{\|\widehat f_\ell(\mat{X}_\ell)-f_\ell(\mat{X}_\ell)\|_F^2}
 {\|f_\ell(\mat{X}_\ell)\|_F^2},
 \label{eq:loss}
\end{equation}
where $f_\ell$ is block $\ell$ in floating point, $\widehat f_\ell$ is the same block with S, GH and the host quantizers at qkv, fc1 and fc2, $\mat{X}_\ell\in\mathbb{R}^{M\times C_\ell}$ is the cached floating-point block input with $M$ tokens and $C_\ell$ channels, and $\|\cdot\|_F$ is the Frobenius norm. We minimize this block objective as BRECQ does~\cite{brecq}; Sec.~\ref{sec:setup} gives the schedule.

\subsection{Placement and Overhead}
\label{sec:rule}
GH runs inside the kernel that produces the layer input, which is the LN kernel for qkv and fc1 and the fc1 GEMM epilogue for fc2 (Fig.~\ref{fig:method}b). S folds into the LN weight and bias at qkv and fc1 and is applied in the fc1 epilogue at fc2, and $\mat{H}^{\top}\mat{D}^{-1}$ folds into the weights. The gain of A folds into the per-channel weight scales and its bias into the layer biases, both applied in the epilogue of the quantized GEMM. The extra computation is therefore the grouped rotation at the three inputs, $O(K\log_2 n)$ additions per token for groups of $n$ channels (Sec.~\ref{sec:group}), and one multiplication by $\mat{D}$ at fc2, both inside the producing kernel. Overall, the three transforms add little arithmetic and no separate pass over memory.

\begin{table*}[t]
\footnotesize
\setlength{\tabcolsep}{1.5pt}
\caption{\textbf{W4A4 results with GH added to four PTQ methods and to AHCQ-SAM.} Left: ImageNet top-1 (\%). Right: SAM mask AP on COCO with DINO~\cite{dinodet} boxes and SAM2.1-T J\&F on SA-V, in simulation. Top: results reported by each method, calibrated on at most 1,024 images. The floating-point row is measured on our checkpoints. Bottom: each method with its official code ($\dagger$), then with GH before its unchanged quantizers; RepQ-ViT also with S and A; AHCQ-SAM with GH on the encoder qkv, fc1 and fc2 only. Small numbers: gain over the $\dagger$ row. Bold: best result per column. Checkpoints and the Swin-B zero-point fix ($^\ast$): Sec.~\ref{supp:imagenet} of the supplement.}
\label{tab:plugin}
\begin{tabular*}{\textwidth}{@{}l@{\extracolsep{\fill}}r@{\extracolsep{0pt}\,}>{\scriptsize\color{black!60}}l@{\extracolsep{\fill}}r@{\extracolsep{0pt}\,}>{\scriptsize\color{black!60}}l@{\extracolsep{\fill}}r@{\extracolsep{0pt}\,}>{\scriptsize\color{black!60}}l@{\extracolsep{\fill}}r@{\extracolsep{0pt}\,}>{\scriptsize\color{black!60}}l@{\extracolsep{\fill}}r@{\extracolsep{0pt}\,}>{\scriptsize\color{black!60}}l@{\extracolsep{\fill}}r@{\extracolsep{0pt}\,}>{\scriptsize\color{black!60}}l@{\extracolsep{\fill}}r@{\extracolsep{0pt}\,}>{\scriptsize\color{black!60}}l@{\extracolsep{\fill}}r@{\extracolsep{0pt}\,}>{\scriptsize\color{black!60}}l@{\extracolsep{\fill}}r@{\extracolsep{0pt}\,}>{\scriptsize\color{black!60}}l@{\extracolsep{\fill}}r@{\extracolsep{0pt}\,}>{\scriptsize\color{black!60}}l@{\extracolsep{\fill}}r@{\extracolsep{0pt}\,}>{\scriptsize\color{black!60}}l@{}}
\toprule
& \multicolumn{14}{c}{ImageNet top-1} & \multicolumn{6}{c}{SAM, mask AP} & \multicolumn{2}{c}{SAM2.1}\\
\cmidrule(lr){2-15}\cmidrule(lr){16-21}\cmidrule(lr){22-23}
Method & \multicolumn{2}{c}{ViT-S} & \multicolumn{2}{c}{ViT-B} & \multicolumn{2}{c}{DeiT-T} & \multicolumn{2}{c}{DeiT-S} & \multicolumn{2}{c}{DeiT-B} & \multicolumn{2}{c}{Swin-S} & \multicolumn{2}{c}{Swin-B} & \multicolumn{2}{c}{B} & \multicolumn{2}{c}{L} & \multicolumn{2}{c}{H} & \multicolumn{2}{c@{}}{T, J\&F}\\
\midrule
Full precision & 81.39 &  & 85.10 &  & 72.21 &  & 79.85 &  & 81.80 &  & 83.32 &  & 85.27 &  & 44.5 &  & 48.6 &  & 49.1 &  & 77.13 & \\
\midrule
RepQ-ViT~\cite{repqvit} & 65.05 &  & 68.48 &  & 57.43 &  & 69.03 &  & 75.61 &  & 79.45 &  & 78.32 &  &  &  &  &  &  &  &  & \\
RepQ-ViT + QwT~\cite{qwt} & 70.8\phantom{0} &  & 76.3\phantom{0} &  & 61.4\phantom{0} &  & 71.5\phantom{0} &  & \multicolumn{1}{c@{\extracolsep{0pt}\,}}{--} &  & 80.4\phantom{0} &  & \multicolumn{1}{c@{\extracolsep{0pt}\,}}{--} &  &  &  &  &  &  &  &  & \\
ERQ~\cite{erqtpami} & 71.61 &  & 78.65 &  & 61.79 &  & 74.35 &  & 79.18 &  & 81.19 &  & 83.32 &  &  &  &  &  &  &  &  & \\
I\&S-ViT~\cite{isvit} & 74.87 &  & 80.07 &  & 65.21 &  & 75.81 &  & 79.97 &  & 81.17 &  & 82.60 &  &  &  &  &  &  &  &  & \\
APHQ-ViT~\cite{aphqvit} & 76.07 &  & 82.41 &  & 66.66 &  & 76.40 &  & 80.21 &  & 81.81 &  & 83.42 &  &  &  &  &  &  &  &  & \\
FIMA-Q~\cite{fimaq} & 76.68 &  & 83.04 &  & 66.84 &  & 76.87 &  & 80.33 &  & 81.82 &  & 83.60 &  &  &  &  &  &  &  &  & \\
LS-ViT~\cite{lsvit} & 76.67 &  & 83.08 &  & 67.05 &  & 76.89 &  & 80.41 &  & 81.82 &  & 83.62 &  &  &  &  &  &  &  &  & \\
AHCPTQ~\cite{ahcptq} &  &  &  &  &  &  &  &  &  &  &  &  &  &  & 16.8 &  & 36.6 &  & 41.2 &  &  & \\
AHCQ-SAM~\cite{ahcqsam} &  &  &  &  &  &  &  &  &  &  &  &  &  &  & 21.2 &  & 40.2 &  & 42.4 &  & 49.71 & \\
\midrule
RepQ-ViT$^\dagger$ & 65.24 &  & 72.13 &  & 57.95 &  & 68.93 &  & 75.68 &  & 79.69 &  & 80.71\rlap{$^\ast$} &  &  &  &  &  &  &  &  & \\
\quad + GH & 72.61 & +7.37 & 81.08 & +8.94 & 62.40 & +4.44 & 74.57 & +5.64 & 79.37 & +3.69 & 81.23 & +1.54 & 82.45 & +1.74 &  &  &  &  &  &  &  & \\
\quad + GH, S, A & 75.73 & +10.49 & 82.33 & +10.20 & 65.03 & +7.08 & 76.05 & +7.12 & 80.19 & +4.51 & 81.48 & +1.80 & 83.19 & +2.48 &  &  &  &  &  &  &  & \\
I\&S-ViT$^\dagger$ & 74.86 &  & 82.40 &  & 65.22 &  & 75.45 &  & 79.98 &  & 81.35 &  & 82.55 &  &  &  &  &  &  &  &  & \\
\quad + GH & 76.05 & +1.19 & 83.10 & +0.70 & 66.94 & +1.72 & 76.94 & +1.49 & 80.55 & +0.57 & 81.45 & +0.10 & 83.04 & +0.48 &  &  &  &  &  &  &  & \\
FIMA-Q$^\dagger$ & 76.34 &  & 82.96 &  & 66.99 &  & 76.68 &  & 80.26 &  & 81.76 &  & 83.50 &  &  &  &  &  &  &  &  & \\
\quad + GH & 77.94 & +1.60 & 83.58 & +0.63 & 68.26 & +1.27 & \textbf{77.52} & +0.84 & 80.69 & +0.43 & \textbf{82.20} & +0.44 & 84.03 & +0.53 &  &  &  &  &  &  &  & \\
LS-ViT$^\dagger$ & 76.45 &  & 83.05 &  & 66.91 &  & 76.72 &  & 80.42 &  & 81.81 &  & 83.54 &  &  &  &  &  &  &  &  & \\
\quad + GH & \textbf{78.08} & +1.64 & \textbf{83.65} & +0.60 & \textbf{68.30} & +1.39 & 77.50 & +0.77 & \textbf{80.86} & +0.44 & 82.11 & +0.29 & \textbf{84.05} & +0.51 &  &  &  &  &  &  &  & \\
AHCQ-SAM$^\dagger$ &  &  &  &  &  &  &  &  &  &  &  &  &  &  & 22.4 &  & 39.4 &  & 43.0 &  & 50.13 & \\
\quad + GH (encoder linears) &  &  &  &  &  &  &  &  &  &  &  &  &  &  & \textbf{28.5} & +6.1 & \textbf{44.4} & +5.0 & \textbf{46.0} & +3.0 & \textbf{54.64} & +4.51\\
\bottomrule
\end{tabular*}
\end{table*}

\section{Experiments}
\label{sec:exp}

\subsection{Setup}
\label{sec:setup}
We implement the grouped Hadamard rotation GH, the channel scaling S and the output affine map A of Sec.~\ref{sec:method} in PyTorch in the official code of each host and in CUDA in our INT4 engine. Unless stated otherwise, GH acts at the inputs of qkv, fc1 and fc2 of every transformer block of the encoder, with groups of $n=64$ at fc2, a width that fits the 128-channel tile of the fc1 epilogue, is the narrowest on the accuracy plateau from 64 to 384 channels (Sec.~\ref{supp:ladder} of the supplement) and is the widest power of two that divides the fc2 width of every model in the paper. At qkv and fc1 the width is $n=128$, except where $K$ is not a multiple of 128, as in DeiT-T and the early Swin-S and Hiera stages, where it is the largest power of two that divides $K$. S and A are used only with RepQ-ViT (Table~\ref{tab:plugin}) and in the SAM2.1 engine (Table~\ref{tab:main}), which have no weight reconstruction and still trail floating point with GH alone (Sec.~\ref{sec:host}). The reconstruction hosts, AHCQ-SAM included, take GH alone. A is fitted with 1,000 AdamW steps per block at a batch size of one image.

\begin{table*}[t]
\footnotesize
\setlength{\tabcolsep}{3pt}
\caption{\textbf{SAM2.1 Hiera encoders}: point-prompt mIoU (W4A4 rows: mean and standard deviation of five calibrations) and encoder latency (ms) on an RTX 4090 at batch 1 and 8. TensorRT INT8: explicit Q/DQ plans calibrated with ModelOpt (per-channel weight scales and one static scale per activation tensor), every GEMM and the windowed attention on INT8 and the rest in FP16, latency of the max plans. TensorRT SC8: the same plans with Q/DQ at qkv, fc1 and fc2 only, the scope of our engine, latency of the plans with the reported calibration. mIoU of both is the best of max, entropy and percentile calibration per encoder.}
\label{tab:main}
\begin{tabular*}{\textwidth}{@{}l@{\extracolsep{\fill}}c ccc ccc ccc ccc@{}}
\toprule
& & \multicolumn{3}{c}{Hiera-T} & \multicolumn{3}{c}{Hiera-S} & \multicolumn{3}{c}{Hiera-B+} & \multicolumn{3}{c}{Hiera-L}\\
\cmidrule(lr){3-5}\cmidrule(lr){6-8}\cmidrule(lr){9-11}\cmidrule(lr){12-14}
Method & W/A & mIoU & b1 (ms) & b8 (ms) & mIoU & b1 (ms) & b8 (ms) & mIoU & b1 (ms) & b8 (ms) & mIoU & b1 (ms) & b8 (ms)\\
\midrule
FP32 reference & 32/32 & 60.68 & -- & -- & 62.11 & -- & -- & 60.41 & -- & -- & 68.70 & -- & --\\
TensorRT FP16 & 16/16 & 60.66 & 3.43 & 41.74 & 62.09 & 3.90 & 46.55 & 60.39 & 6.03 & 73.81 & 68.72 & 14.32 & 152.57\\
TensorRT INT8 & 8/8 & 56.96 & 3.40 & 30.52 & 58.38 & 3.94 & 34.39 & 58.61 & 7.07 & 64.80 & 55.84 & 12.92 & 117.30\\
TensorRT SC8 & 8/8 & 60.02 & 3.50 & 38.80 & 61.66 & 4.06 & 43.34 & 60.40 & 6.49 & 72.94 & 67.63 & 13.81 & 145.06\\
Ours W8A8 & 8/8 & 60.71 & 2.19 & 16.81 & 62.15 & 2.61 & 19.30 & 60.40 & 4.13 & 32.72 & 68.63 & 9.09 & 72.40\\
GH & 4/4 & 56.57$\pm$0.58 & 1.86 & 14.92 & 58.02$\pm$0.50 & 2.18 & 16.78 & 58.51$\pm$0.33 & 3.37 & 27.22 & 64.59$\pm$0.27 & 7.09 & 57.05\\
SGHAQuant & 4/4 & 60.74$\pm$0.12 & 1.87 & 15.01 & 62.40$\pm$0.19 & 2.19 & 16.90 & 60.62$\pm$0.17 & 3.38 & 27.47 & 68.34$\pm$0.18 & 7.17 & 57.53\\
\bottomrule
\end{tabular*}
\end{table*}

\subsection{Rotation as a Plug-in}
\label{sec:plugin}

We add GH as a plug-in to RepQ-ViT~\cite{repqvit}, I\&S-ViT~\cite{isvit}, FIMA-Q~\cite{fimaq} and LS-ViT~\cite{lsvit} on ImageNet, where we report validation top-1, and to AHCQ-SAM~\cite{ahcqsam} on SAM ViT-B/L/H and SAM2.1 Tiny. Each host keeps its quantization scope, quantizers and calibration schedule unchanged, and on the ImageNet hosts GH replaces the LN reparameterization at qkv and fc1. The exception is AHCQ-SAM. Its encoder qkv, fc1 and fc2 take the quantizers of Sec.~\ref{sec:samv1}, fitted beforehand and frozen, and its calibration and reconstruction cover the rest (Sec.~\ref{supp:imagenet} of the supplement). The S statistics and the A fit on RepQ-ViT use 512 further training images. The other ImageNet hosts retain their original weight reconstruction procedures, using 1,024 training images. Table~\ref{tab:plugin} reports the plug-in results. GH improves all 28 method-model pairs. The gains are 1.5 to 8.9 points on RepQ-ViT and 0.1 to 1.7 points on the reconstruction hosts. For the eight pairs with the smallest gains, three calibration draws give mean GH gains of $+0.18$ to $+0.60$ points with standard deviations of at most 0.10. The gain is positive in every draw, and the 95\% image-bootstrap confidence interval for the mean gain excludes zero for every pair (Sec.~\ref{supp:imagenet} of the supplement). With GH, S and A, RepQ-ViT, which has no weight reconstruction, comes within 2 points of the reconstruction hosts without GH on every model.
On AHCQ-SAM (right block of Table~\ref{tab:plugin}), mask AP rises by 6.1, 5.0 and 3.0 points on SAM ViT-B, ViT-L and ViT-H, and video J\&F rises by 4.5 points on SAM2.1 Tiny.

\subsection{SAM2.1 Encoders on the Native INT4 Engine}
\label{sec:sam21}

On the four SAM2.1 Hiera encoders~\cite{sam2,hiera}, qkv, fc1 and fc2 run at W4A4 on INT4 tensor cores, with round-to-nearest weights at one scale per output channel, per-token activation scales at qkv and fc1 and one static scale per layer at fc2. The output projection, the attention matrix products, the patch embedding and the neck run in FP16 in the engine; the prompt encoder and the mask decoder run in FP32. We report point-prompt mIoU, the per-instance mask IoU averaged over 4,629 instances of the first 1,000 COCO~\cite{coco} validation images with one interior point per instance, measured on the outputs of our native INT4 executable after calibration on 32 COCO training images. Latencies are measured on the same machine equipped with an RTX 4090, using CUDA graphs (encoder and neck, batch sizes 1 and 8). W8A8 is the 8-bit path of the same engine, with the same grouped rotation and GPTQ weights~\cite{gptq}. TensorRT 11.2 runs FP16 plans and explicit INT8 plans calibrated with ModelOpt.

Table~\ref{tab:main} reports mIoU and latency on the engine. With GH alone, the four encoders lose 1.9 to 4.1 mIoU against floating point. With S and A added, they are within 0.4 mIoU of floating point. On the RTX 4090, the W4A4 encoders run 1.17 to 1.27$\times$ (batch 1) and 1.12 to 1.26$\times$ (batch 8) faster than the W8A8 path of the same engine, and 1.78 to 2.00$\times$ (batch 1) and 2.65 to 2.78$\times$ (batch 8) faster than TensorRT FP16. TensorRT SC8, the INT8 plan at the quantization scope of our engine, is within 1.1 mIoU of TensorRT FP16, and the W4A4 encoders run 1.84 to 1.91$\times$ (batch 1) and 2.51 to 2.65$\times$ (batch 8) faster than it. Sec.~\ref{supp:runtime} of the supplement repeats the measurement on an RTX 3090, where they run 1.13 to 1.30$\times$ (batch 1) and 1.15 to 1.30$\times$ (batch 8) faster than W8A8, and 1.86 to 2.00$\times$ (batch 1) and 2.20 to 2.25$\times$ (batch 8) faster than TensorRT FP16. Computing S and fitting A take 2 to 9 minutes per encoder on one RTX 4090 (Hiera-T to Hiera-L), excluding the one-time calibration forward passes.

\begin{table}[t]
\footnotesize
\setlength{\tabcolsep}{3pt}
\caption{\textbf{Run-time cost of the transforms}, percent of the W4A4 encoder latency of Table~\ref{tab:main} at batch 1, measured with each transform switched off. Full-width: a randomized full-width Hadamard at fc2 in a separate kernel; its accuracy row is simulated with the signed full-width rotation at all three inputs.}
\label{tab:cost}
\begin{tabular*}{\columnwidth}{@{}l@{\extracolsep{\fill}}rrrr@{}}
\toprule
Piece (\% of W4A4 latency) & T & S & B+ & L\\
\midrule
GH at qkv, fc1 (LN kernel) & 2.25 & 2.55 & 1.83 & 2.79\\
GH at fc2 (fc1 epilogue) & 0.66 & 0.70 & 1.04 & 2.30\\
GH, all three & 2.91 & 3.24 & 2.86 & 5.09\\
S and A (one multiply) & 0.63 & 0.73 & 0.53 & 1.18\\
\midrule
randomized full-width fc2 & +10.5 & +10.6 & +27.3 & +20.1\\
\quad mIoU vs.\ GH, S, A (sim.) & $-0.01$ & $-0.30$ & $+0.08$ & $-0.24$\\
\bottomrule
\end{tabular*}
\end{table}

\paragraph{Run-time cost.}
\label{sec:cost}
Table~\ref{tab:cost} reports the cost of each transform. GH costs 2.9 to 5.1\% of the W4A4 encoder latency and S and A together 0.5 to 1.2\%. A full-width Hadamard at fc2 needs a separate kernel that adds 10 to 27\% and cuts the W4A4 advantage over W8A8 at batch 1 from 1.17 to 1.27$\times$ to 0.96 to 1.08$\times$. The pass is memory-bound. On Hiera-L at batch 1 it reads and writes 1.37 GB in 1.55 ms, 885 GB/s against the 1,008 GB/s of the RTX 4090, so its cost is a lower bound for every transform over the full fc2 row. In simulation, the signed full-width rotation at all three inputs is within 0.3 points of GH on all four encoders ($-0.30$ to $+0.08$), with all paired 95\% confidence intervals including zero (see Sec.~\ref{supp:sam21} of the supplement for the image-bootstrap procedure). With S and A, the signed full-width rotation thus matches GH in accuracy but requires an additional pass over the fc2 activations.

\subsection{GH on the SAM Encoders}
\label{sec:samv1}

\begin{table}[t]
\footnotesize
\setlength{\tabcolsep}{2.4pt}
\caption{\textbf{GH on the SAM ViT encoders at the full encoder coverage of PTQ4SAM}: every encoder operand at W4A4 (qkv, proj, fc1, fc2, the attention operands and the neck) with round-to-nearest weights, without GH (none) and with GH at every GEMM input (qkv, proj, fc1, fc2 and the neck); both rows keep a per-head Hadamard on Q, K and V; decoder in floating point, simulation. The fc2 activation scale is static per layer. COCO mask AP with YOLOX~\cite{yolox} and DINO boxes.}
\label{tab:samv1}
\centering
\begin{tabular}{l ccc ccc}
\toprule
& \multicolumn{3}{c}{YOLOX boxes (mask AP)} & \multicolumn{3}{c}{DINO boxes (mask AP)}\\
\cmidrule(lr){2-4}\cmidrule(lr){5-7}
& ViT-B & ViT-L & ViT-H & ViT-B & ViT-L & ViT-H\\
\midrule
Floating point & 37.2 & 40.4 & 41.0 & 44.5 & 48.6 & 49.1\\
W4A4, none & 33.2 & 37.8 & 38.2 & 39.2 & 45.0 & 45.2\\
W4A4, GH & 36.3 & 40.0 & 40.6 & 43.4 & 47.9 & 48.6\\
\bottomrule
\end{tabular}
\end{table}

Table~\ref{tab:samv1} tests GH on the SAM ViT-B, ViT-L and ViT-H encoders at the full encoder coverage of PTQ4SAM, with qkv, proj, fc1, fc2, the attention operands and the neck at W4A4 and the decoder in floating point. Weights are rounded to nearest. At qkv, fc1 and fc2, the clipping ratio of each weight row and of the activation quantizer is learned per block with the loss of Eq.~\ref{eq:loss}, as in the learned weight clipping of OmniQuant~\cite{omniquant}. S and A are not used. GH alone keeps the three encoders within 1.1 AP of floating point with YOLOX and DINO boxes; without GH they lose 2.6 to 5.3 AP, and the paired gain of GH is positive in every cell. Under a harder quantization scope and a different encoder family, these rows further verify the effectiveness of GH.
\subsection{GH on DINOv3}
\label{sec:dinov3}
\begin{table}[t]
\footnotesize
\setlength{\tabcolsep}{4pt}
\caption{\textbf{GH on DINOv3-L}: ImageNet zero-shot top-1 (\%) and encoder latency (ms) on an RTX 4090 at batch 1, on our engine. Ours W8A8: the 8-bit path of Sec.~\ref{sec:sam21}, with GH and GPTQ weights. W4A4 rows: round-to-nearest weights at qkv, fc1 and fc2, per-token scales at qkv and fc1 and one static fc2 scale per layer, without rotation (none), with the full-width Hadamard (a separate pass at fc2) and with GH. The fc2 of block 0, proj and the attention products stay in floating point in every row.}
\label{tab:dinov3}
\centering
\begin{tabular}{l cc ccc}
\toprule
& & & \multicolumn{3}{c}{W4A4}\\
\cmidrule(lr){4-6}
& FP32 & Ours W8A8 & none & full-width H & GH\\
\midrule
top-1 & 81.39 & 81.39 & 0.48 & 0.10 & 80.44\\
b1 (ms) & -- & 1.27 & 0.93 & 1.18 & 0.97\\
\bottomrule
\end{tabular}
\end{table}
Table~\ref{tab:dinov3} tests GH on DINOv3-L~\cite{dinov3} (ViT-L/16 at 224 px) on our engine, with the ImageNet zero-shot top-1 of its text-aligned head (dino.txt) on the 50,000 validation images as the metric. The qkv, fc1 and fc2 layers of each block use W4A4 quantization with round-to-nearest weights, per-token activation scales at qkv and fc1, and one static activation scale per fc2 layer calibrated on 64 COCO images. The exception is fc2 in block 0, whose static range is set by one token per image; this layer and all remaining operators stay in floating point in both the W4A4 and W8A8 paths. No S or A is used. The W8A8 path, with the same GH, matches floating point. At W4A4, GH loses 0.95 top-1 against floating point, and the rows without rotation or with the full-width rotation collapse below 1\% top-1. The W4A4 path is a further 1.30$\times$ faster than W8A8 at batch 1. On a third encoder family, the result again shows the instability of the full-width rotation and the effectiveness of the grouped rotation.
\subsection{Ablation on SAM2.1}
\label{sec:ablation}
\begin{table}[t]
\footnotesize
\setlength{\tabcolsep}{3.5pt}
\caption{\textbf{Ablation on SAM2.1}, point-prompt mIoU on the engine, mean of two fitting seeds. (a)~none: no GH, S or A, in simulation (the simulated GH row is within 0.2 of the engine row); then each addition relative to GH; the last row removes GH. (b)~A per-layer closed-form affine (our reimplementation of QwT-v2~\cite{qwtv2}) in place of A, as a difference to SGHAQuant. Both corrections use the same calibration inputs and layer set, with identical S, GH and static fc2 scales. $*$: the 95\% confidence interval from a paired bootstrap over the images excludes 0. (c)~Weights rounded to nearest or by GPTQ, each without and with A; mean of two seeds, simulation.}
\label{tab:ablation}
\begin{tabular*}{\columnwidth}{@{}l@{\extracolsep{\fill}}cccc@{}}
\toprule
& T & S & B+ & L\\
\midrule
FP32 reference & 60.68 & 62.11 & 60.41 & 68.70\\
\midrule
\multicolumn{5}{@{}l}{\emph{(a) parts}}\\
none (sim.) & 50.79 & 54.34 & 55.95 & 57.57\\
GH & 56.10 & 58.18 & 58.61 & 64.40\\
+ S & 58.04 & 59.28 & 58.90 & 66.38\\
+ A & 59.35 & 61.77 & 60.67 & 67.71\\
+ S and A (SGHAQuant) & \textbf{60.78} & \textbf{62.42} & 60.57 & \textbf{68.38}\\
S and A, no GH & 58.17 & 60.12 & \textbf{60.78} & 65.00\\
\midrule
\multicolumn{5}{@{}l}{\emph{(b) A replaced by a per-layer closed form, vs.\ SGHAQuant}}\\
QwT-v2 closed form~\cite{qwtv2} & $-1.37^*$ & $-1.33^*$ & $-0.19$ & $-0.41^*$\\
\midrule
\multicolumn{5}{@{}l}{\emph{(c) weights after GH and S, simulation (fit minutes, T to L)}}\\
round to nearest & 58.14 & 59.17 & 59.07 & 66.40\\
GPTQ~\cite{gptq} (0.3 to 2.1) & 58.68 & 59.78 & 59.01 & 65.87\\
round to nearest, A (1.9 to 9.2) & 60.83 & 62.54 & 60.74 & 68.48\\
GPTQ, A & 60.86 & 62.29 & 60.71 & 68.46\\
\bottomrule
\end{tabular*}
\end{table}

Table~\ref{tab:ablation}a adds the parts one at a time. S and A each help alone, A helps more than S on every encoder, and their gains are nearly additive on the T, S and L encoders. Removing GH from SGHAQuant costs 2.3 to 3.4 points on T, S and L, with 95\% confidence intervals from a paired bootstrap over the images excluding zero, and changes B+ by $+0.21$ within its interval. Replacing A by the per-layer closed form of QwT-v2~\cite{qwtv2} (Table~\ref{tab:ablation}b) costs 0.2 to 1.4 points, with intervals excluding zero on T, S and L. Against GPTQ, which is in the same time class, A adds 1.7 to 3.4 points over rounding to nearest where GPTQ changes mIoU by $-0.5$ to $+0.6$, and A on GPTQ weights is within 0.3 points of A on rounded weights (Table~\ref{tab:ablation}c). On SAM2.1 at least, A thus gains more accuracy than the closed form of QwT-v2 and than GPTQ, and GPTQ adds little on top of A.

\section{Conclusion}
The unsigned Hadamard rotation is unstable on vision transformers at the GELU-fed fc2 input, where the all-ones row collects each token's channel sum into one coordinate. Groups of 64 channels bound this coordinate inside the kernel that produces the activation. Inserted before the quantizers of four PTQ methods, the grouped rotation improves all 28 method-model pairs on seven ImageNet models. With SGHAQuant, our native INT4 SAM2.1 encoders stay within 0.4 mIoU of floating point and run 1.12 to 1.27$\times$ faster than the same engine's W8A8 path.

\paragraph{Limitations.}
Every cell of Table~\ref{tab:plugin} is one calibration draw. The engine quantizes qkv, fc1 and fc2 only, and the W4A4 speedup needs INT4 tensor cores and is measured on one RTX 4090 and one RTX 3090; the Hopper and Blackwell generations drop INT4 tensor cores, so the speedup applies to the installed Ampere and Ada GPUs. Table~\ref{tab:samv1} and the AHCQ-SAM rows of Table~\ref{tab:plugin} are simulations; the SAM mask decoder is out of scope except inside AHCQ-SAM. Without reconstruction, the full-width Hadamard with random signs is up to one point above GH (Table~\ref{tab:ladder}), at the cost of the separate fc2 pass of Sec.~\ref{sec:cost}; should that pass become cheap, the randomized full-width rotation would be a better choice, whereas with S and A the two are within 0.3 points.

\section*{LLM Usage Statement}
We used Claude Code (Anthropic), an LLM-based coding assistant, to write and refactor code under our direction, to run experiments and collect their logs, to search the literature, and to edit the writing. It generated no data or results; the method, the experiments, the baselines and the claims are ours, and we verified every reported number and take responsibility for the content.

{\small
\bibliographystyle{ieeenat_fullname}
\bibliography{references}

@inproceedings{sam2,
  title={{SAM 2}: Segment Anything in Images and Videos},
  author={Ravi, Nikhila and Gabeur, Valentin and Hu, Yuan-Ting and Hu, Ronghang and Ryali, Chaitanya and Ma, Tengyu and Khedr, Haitham and R{\"a}dle, Roman and Rolland, Chloe and Gustafson, Laura and others},
  booktitle={ICLR},
  year={2025}
}

@inproceedings{hiera,
  title={Hiera: A Hierarchical Vision Transformer without the Bells-and-Whistles},
  author={Ryali, Chaitanya and Hu, Yuan-Ting and Bolya, Daniel and Wei, Chen and Fan, Haoqi and Huang, Po-Yao and Aggarwal, Vaibhav and Chowdhury, Arkabandhu and Poursaeed, Omid and Hoffman, Judy and others},
  booktitle={ICML},
  year={2023}
}

@inproceedings{gptq,
  title={{GPTQ}: Accurate Post-Training Quantization for Generative Pre-trained Transformers},
  author={Frantar, Elias and Ashkboos, Saleh and Hoefler, Torsten and Alistarh, Dan},
  booktitle={ICLR},
  year={2023}
}

@inproceedings{smoothquant,
  title={{SmoothQuant}: Accurate and Efficient Post-Training Quantization for Large Language Models},
  author={Xiao, Guangxuan and Lin, Ji and Seznec, Mickael and Wu, Hao and Demouth, Julien and Han, Song},
  booktitle={ICML},
  year={2023}
}

@inproceedings{quarot,
  title={{QuaRot}: Outlier-Free 4-Bit Inference in Rotated {LLMs}},
  author={Ashkboos, Saleh and Mohtashami, Amirkeivan and Croci, Maximilian L. and Li, Bo and Cameron, Pashmina and Jaggi, Martin and Alistarh, Dan and Hoefler, Torsten and Hensman, James},
  booktitle={NeurIPS},
  year={2024}
}

@inproceedings{spinquant,
  title={{SpinQuant}: {LLM} Quantization with Learned Rotations},
  author={Liu, Zechun and Zhao, Changsheng and Fedorov, Igor and Soran, Bilge and Choudhary, Dhruv and Krishnamoorthi, Raghuraman and Chandra, Vikas and Tian, Yuandong and Blankevoort, Tijmen},
  booktitle={ICLR},
  year={2025}
}

@inproceedings{qserve,
  title={{QServe}: {W4A8KV4} Quantization and System Co-design for Efficient {LLM} Serving},
  author={Lin, Yujun and Tang, Haotian and Yang, Shang and Zhang, Zhekai and Xiao, Guangxuan and Gan, Chuang and Han, Song},
  booktitle={MLSys},
  year={2025}
}

@inproceedings{ptq4vit,
  title={{PTQ4ViT}: Post-Training Quantization for Vision Transformers with Twin Uniform Quantization},
  author={Yuan, Zhihang and Xue, Chenhao and Chen, Yiqi and Wu, Qiang and Sun, Guangyu},
  booktitle={ECCV},
  year={2022}
}

@inproceedings{repqvit,
  title={{RepQ-ViT}: Scale Reparameterization for Post-Training Quantization of Vision Transformers},
  author={Li, Zhikai and Xiao, Junrui and Yang, Lianwei and Gu, Qingyi},
  booktitle={ICCV},
  pages={17227--17236},
  year={2023}
}

@inproceedings{adalog,
  title={{AdaLog}: Post-Training Quantization for Vision Transformers with Adaptive Logarithm Quantizer},
  author={Wu, Zhuguanyu and Chen, Jiaxin and Zhong, Hanwen and Huang, Di and Wang, Yunhong},
  booktitle={ECCV},
  year={2024}
}

@inproceedings{ptq4sam,
  title={{PTQ4SAM}: Post-Training Quantization for Segment Anything},
  author={Lv, Chengtao and Chen, Hong and Guo, Jinyang and Ding, Yifu and Liu, Xianglong},
  booktitle={CVPR},
  year={2024}
}

@inproceedings{vit,
  title={An Image is Worth 16x16 Words: Transformers for Image Recognition at Scale},
  author={Dosovitskiy, Alexey and Beyer, Lucas and Kolesnikov, Alexander and Weissenborn, Dirk and Zhai, Xiaohua and Unterthiner, Thomas and Dehghani, Mostafa and Minderer, Matthias and Heigold, Georg and Gelly, Sylvain and Uszkoreit, Jakob and Houlsby, Neil},
  booktitle={ICLR},
  year={2021}
}

@article{dinov3,
  title={{DINOv3}},
  author={Sim{\'e}oni, Oriane and Vo, Huy V. and Seitzer, Maximilian and Baldassarre, Federico and Oquab, Maxime and Jose, Cijo and Khalidov, Vasil and Szafraniec, Marc and Yi, Seungeun and Ramamonjisoa, Micha{\"e}l and others},
  journal={arXiv preprint arXiv:2508.10104},
  year={2025}
}

@inproceedings{flatquant,
  title={{FlatQuant}: Flatness Matters for {LLM} Quantization},
  author={Sun, Yuxuan and Liu, Ruikang and Bai, Haoli and Bao, Han and Zhao, Kang and Li, Yuening and Hu, Jiaxin and Yu, Xianzhi and Hou, Lu and Yuan, Chun and Jiang, Xin and Liu, Wulong and Yao, Jun},
  booktitle={ICML},
  year={2025}
}

@inproceedings{brecq,
  title={{BRECQ}: Pushing the Limit of Post-Training Quantization by Block Reconstruction},
  author={Li, Yuhang and Gong, Ruihao and Tan, Xu and Yang, Yang and Hu, Peng and Zhang, Qi and Yu, Fengwei and Wang, Wei and Gu, Shi},
  booktitle={ICLR},
  year={2021}
}

@inproceedings{sam,
  title={Segment Anything},
  author={Kirillov, Alexander and Mintun, Eric and Ravi, Nikhila and Mao, Hanzi and Rolland, Chloe and Gustafson, Laura and Xiao, Tete and Whitehead, Spencer and Berg, Alexander C. and Lo, Wan-Yen and Doll{\'a}r, Piotr and Girshick, Ross},
  booktitle={ICCV},
  year={2023}
}

@inproceedings{svdquant,
  title={{SVDQuant}: Absorbing Outliers by Low-Rank Components for 4-Bit Diffusion Models},
  author={Li, Muyang and Lin, Yujun and Zhang, Zhekai and Cai, Tianle and Li, Xiuyu and Guo, Junxian and Xie, Enze and Meng, Chenlin and Zhu, Jun-Yan and Han, Song},
  booktitle={ICLR},
  year={2025}
}

@inproceedings{qattn,
  title={{QAttn}: Efficient {GPU} Kernels for Mixed-Precision Vision Transformers},
  author={Kluska, Piotr and Castell{\'o}, Adri{\'a}n and Scheidegger, Florian and Malossi, A. Cristiano I. and Quintana-Ort{\'i}, Enrique S.},
  booktitle={CVPR Workshops},
  pages={3648--3657},
  year={2024}
}

@article{ahcqsam,
  title={{AHCQ-SAM}: Toward Accurate and Hardware-Compatible Post-Training Segment Anything Model Quantization},
  author={Zhang, Wenlun and Zhong, Yunshan and Yan, Weiqi and Zhang, Shengchuan and Ando, Shimpei and Yoshioka, Kentaro},
  journal={arXiv preprint arXiv:2503.03088},
  year={2026},
  url={https://arxiv.org/abs/2503.03088v4}
}

@article{hadanorm,
  title={{HadaNorm}: Diffusion Transformer Quantization through Mean-Centered Transformations},
  author={Federici, Marco and Del Chiaro, Riccardo and van Breugel, Boris and Whatmough, Paul and Nagel, Markus},
  journal={arXiv preprint arXiv:2506.09932},
  year={2025}
}

@inproceedings{duquant,
  title={{DuQuant}: Distributing Outliers via Dual Transformation Makes Stronger Quantized {LLMs}},
  author={Lin, Haokun and Xu, Haobo and Wu, Yichen and Cui, Jingzhi and Zhang, Yingtao and Mou, Linzhan and Song, Linqi and Sun, Zhenan and Wei, Ying},
  booktitle={NeurIPS},
  year={2024}
}

@inproceedings{fqvit,
  title={{FQ-ViT}: Post-Training Quantization for Fully Quantized Vision Transformer},
  author={Lin, Yang and Zhang, Tianyu and Sun, Peiqin and Li, Zheng and Zhou, Shuchang},
  booktitle={IJCAI},
  year={2022}
}

@inproceedings{omniquant,
  title     = {{OmniQuant}: Omnidirectionally Calibrated Quantization for Large Language Models},
  author    = {Shao, Wenqi and Chen, Mengzhao and Zhang, Zhaoyang and Xu, Peng and Zhao, Lirui and Li, Zhiqian and Zhang, Kaipeng and Gao, Peng and Qiao, Yu and Luo, Ping},
  booktitle = {ICLR},
  year      = {2024}
}

@inproceedings{qdrop,
  title     = {{QDrop}: Randomly Dropping Quantization for Extremely Low-bit Post-Training Quantization},
  author    = {Wei, Xiuying and Gong, Ruihao and Li, Yuhang and Liu, Xianglong and Yu, Fengwei},
  booktitle = {ICLR},
  year      = {2022}
}

@inproceedings{coco,
  title     = {{Microsoft COCO}: Common Objects in Context},
  author    = {Lin, Tsung-Yi and Maire, Michael and Belongie, Serge and Hays, James and Perona, Pietro and Ramanan, Deva and Doll{\'a}r, Piotr and Zitnick, C. Lawrence},
  booktitle = {ECCV},
  pages     = {740--755},
  year      = {2014}
}

@article{yolox,
  title   = {{YOLOX}: Exceeding {YOLO} Series in 2021},
  author  = {Ge, Zheng and Liu, Songtao and Wang, Feng and Li, Zeming and Sun, Jian},
  journal = {arXiv preprint arXiv:2107.08430},
  year    = {2021}
}

@inproceedings{dinodet,
  title     = {{DINO}: {DETR} with Improved DeNoising Anchor Boxes for End-to-End Object Detection},
  author    = {Zhang, Hao and Li, Feng and Liu, Shilong and Zhang, Lei and Su, Hang and Zhu, Jun and Ni, Lionel M. and Shum, Heung-Yeung},
  booktitle = {ICLR},
  year      = {2023}
}

@article{isvit,
  title   = {{I\&S-ViT}: An Inclusive \& Stable Method for Post-Training {ViTs} Quantization},
  author  = {Zhong, Yunshan and Hu, Jiawei and Lin, Mingbao and Chen, Mengzhao and Ji, Rongrong},
  journal = {IEEE Transactions on Pattern Analysis and Machine Intelligence},
  volume  = {48},
  number  = {2},
  pages   = {1063--1080},
  year    = {2026},
  doi     = {10.1109/TPAMI.2025.3610466}
}

@inproceedings{erq,
  title     = {{ERQ}: Error Reduction for Post-Training Quantization of Vision Transformers},
  author    = {Zhong, Yunshan and Hu, Jiawei and Huang, You and Zhang, Yuxin and Ji, Rongrong},
  booktitle = {ICML},
  pages     = {61664--61680},
  year      = {2024}
}

@inproceedings{llmint8,
  title     = {{LLM.int8()}: 8-bit Matrix Multiplication for Transformers at Scale},
  author    = {Dettmers, Tim and Lewis, Mike and Belkada, Younes and Zettlemoyer, Luke},
  booktitle = {NeurIPS},
  pages     = {30318--30332},
  year      = {2022}
}

@inproceedings{outliersupp,
  title     = {{Outlier Suppression}: Pushing the Limit of Low-bit Transformer Language Models},
  author    = {Wei, Xiuying and Zhang, Yunchen and Zhang, Xiangguo and Gong, Ruihao and Zhang, Shanghang and Zhang, Qi and Yu, Fengwei and Liu, Xianglong},
  booktitle = {NeurIPS},
  pages     = {17402--17414},
  year      = {2022}
}

@inproceedings{quip,
  title     = {{QuIP}: 2-Bit Quantization of Large Language Models With Guarantees},
  author    = {Chee, Jerry and Cai, Yaohui and Kuleshov, Volodymyr and De Sa, Christopher},
  booktitle = {NeurIPS},
  pages     = {4396--4429},
  year      = {2023}
}

@inproceedings{quipsharp,
  title     = {{QuIP\#}: Even Better {LLM} Quantization with {Hadamard} Incoherence and Lattice Codebooks},
  author    = {Tseng, Albert and Chee, Jerry and Sun, Qingyao and Kuleshov, Volodymyr and De Sa, Christopher},
  booktitle = {ICML},
  pages     = {48630--48656},
  year      = {2024}
}

@inproceedings{atom,
  title     = {{Atom}: Low-Bit Quantization for Efficient and Accurate {LLM} Serving},
  author    = {Zhao, Yilong and Lin, Chien-Yu and Zhu, Kan and Ye, Zihao and Chen, Lequn and Zheng, Size and Ceze, Luis and Krishnamurthy, Arvind and Chen, Tianqi and Kasikci, Baris},
  booktitle = {MLSys},
  pages     = {196--209},
  year      = {2024}
}

@inproceedings{qwt,
  title     = {Quantization without Tears},
  author    = {Fu, Minghao and Yu, Hao and Shao, Jie and Zhou, Junjie and Zhu, Ke and Wu, Jianxin},
  booktitle = {CVPR},
  pages     = {4462--4472},
  year      = {2025}
}

@article{qwtv2,
  title   = {{QwT-v2}: Practical, Effective and Efficient Post-Training Quantization},
  author  = {Tang, Ningyuan and Fu, Minghao and Yu, Hao and Wu, Jianxin},
  journal = {arXiv preprint arXiv:2505.20932},
  year    = {2025}
}

@inproceedings{aphqvit,
  title     = {{APHQ-ViT}: Post-Training Quantization with Average Perturbation Hessian Based Reconstruction for Vision Transformers},
  author    = {Wu, Zhuguanyu and Zhang, Jiayi and Chen, Jiaxin and Guo, Jinyang and Huang, Di and Wang, Yunhong},
  booktitle = {CVPR},
  pages     = {9686--9695},
  year      = {2025}
}

@inproceedings{fimaq,
  title     = {{FIMA-Q}: Post-Training Quantization for Vision Transformers by Fisher Information Matrix Approximation},
  author    = {Wu, Zhuguanyu and Wang, Shihe and Zhang, Jiayi and Chen, Jiaxin and Wang, Yunhong},
  booktitle = {CVPR},
  pages     = {14891--14900},
  year      = {2025}
}

@inproceedings{lsvit,
  title     = {{LS-ViT}: Least-Squares Hessian Based Block Reconstruction for Low-Bit Post-Training Quantization of Vision Transformers},
  author    = {Hwang, Hyunha and Nguyen, Xuan Truong and Lee, Hyuk-Jae},
  booktitle = {CVPR},
  pages     = {33588--33597},
  year      = {2026}
}

@inproceedings{ahcptq,
  title     = {{AHCPTQ}: Accurate and Hardware-Compatible Post-Training Quantization for Segment Anything Model},
  author    = {Zhang, Wenlun and Zhong, Yunshan and Ando, Shimpei and Yoshioka, Kentaro},
  booktitle = {ICCV},
  pages     = {22383--22392},
  year      = {2025}
}

@inproceedings{carsam,
  title     = {{CAR-SAM}: Cross-Attention Reconstruction for Post-Training Quantization of the Segment Anything Model},
  author    = {Wen, Houji and Yu, Jiangyong and Yang, Dawei and Li, Jun},
  booktitle = {CVPR},
  pages     = {33632--33641},
  year      = {2026}
}

@inproceedings{regcache,
  title     = {Activation Quantization of Vision Encoders Needs Prefixing Registers},
  author    = {Kim, Seunghyeon and Yeom, Taesun and Kim, Jinho and Park, Wonpyo and Kim, Kyuyeun and Lee, Jaeho},
  booktitle = {ECCV},
  pages     = {20--37},
  year      = {2026}
}

@article{erqtpami,
  title   = {Towards Accurate Post-Training Quantization of Vision Transformers via Error Reduction},
  author  = {Zhong, Yunshan and Huang, You and Hu, Jiawei and Zhang, Yuxin and Ji, Rongrong},
  journal = {IEEE Transactions on Pattern Analysis and Machine Intelligence},
  volume  = {47},
  number  = {4},
  pages   = {2676--2692},
  year    = {2025}
}

@article{convrot,
  title   = {{ConvRot}: Rotation-Based Plug-and-Play 4-bit Quantization for Diffusion Transformers},
  author  = {Huang, Feice and Han, Zuliang and Zhou, Xing and Chen, Yihuang and Zhu, Lifei and Wang, Haoqian},
  journal = {arXiv preprint arXiv:2512.03673},
  year    = {2025}
}

@article{parklee2026rotate,
  title   = {When and How to Rotate: Loss Design and Activation Granularity for Rotation-Based Vision Model Quantization},
  author  = {Park, Jaewoo and Lee, Jihae},
  journal = {IEEE Access},
  volume  = {14},
  pages   = {150171--150191},
  year    = {2026},
  doi     = {10.1109/ACCESS.2026.3737447}
}

@inproceedings{brq,
  title     = {Block Rotation is All You Need for {MXFP4} Quantization},
  author    = {Shao, Yuantian and Wang, Peisong and Chen, Yuanteng and Xu, Chang and Wei, Zhihui and Cheng, Jian},
  booktitle = {ICML},
  year      = {2026}
}

@article{centriq,
  title   = {{CentriQ}: Calibration-Free Quantization of Diffusion Transformers via Exact Mean Centering},
  author  = {Jovanovi{\'c}, Nata{\v{s}}a and Salzmann, Mathieu and Javed, Saqib},
  journal = {arXiv preprint arXiv:2610.06260},
  year    = {2026}
}

@article{jointptq,
  title   = {Joint Post-Training Quantization of Vision Transformers with Learned Prompt-Guided Data Generation},
  author  = {Li, Shile and Karmann, Markus and Urfalioglu, Onay},
  journal = {arXiv preprint arXiv:2602.18861},
  year    = {2026}
}

@inproceedings{quantvggt,
  title     = {Quantized Visual Geometry Grounded Transformer},
  author    = {Feng, Weilun and Qin, Haotong and Wu, Mingqiang and Yang, Chuanguang and Li, Yuqi and Li, Xiangqi and An, Zhulin and Huang, Libo and Zhang, Yulun and Magno, Michele and Xu, Yongjun},
  booktitle = {ICLR},
  year      = {2026}
}

@article{sparsesam,
  title   = {{SparseSAM}: Structured Sparsification of Activations in Segment Anything Models},
  author  = {Tran, Hoai-Chau and Nguyen, Chi H. and Nguyen, Duy M. H. and Niepert, Mathias and Lai, Fan and Doan, Khoa D.},
  journal = {arXiv preprint arXiv:2605.17633},
  year    = {2026}
}

@inproceedings{viditq,
  title     = {{ViDiT-Q}: Efficient and Accurate Quantization of Diffusion Transformers for Image and Video Generation},
  author    = {Zhao, Tianchen and Fang, Tongcheng and Huang, Haofeng and Wan, Rui and Soedarmadji, Widyadewi and Liu, Enshu and Li, Shiyao and Lin, Zinan and Dai, Guohao and Yan, Shengen and Yang, Huazhong and Ning, Xuefei and Wang, Yu},
  booktitle = {ICLR},
  year      = {2025}
}

@inproceedings{igqvit,
  title     = {Instance-Aware Group Quantization for Vision Transformers},
  author    = {Moon, Jaehyeon and Kim, Dohyung and Cheon, Junyong and Ham, Bumsub},
  booktitle = {CVPR},
  pages     = {16132--16141},
  year      = {2024}
}

@article{gelu,
  title={Gaussian Error Linear Units ({GELU}s)},
  author={Hendrycks, Dan and Gimpel, Kevin},
  journal={arXiv preprint arXiv:1606.08415},
  year={2016}
}

@article{bengio2013ste,
  title   = {Estimating or Propagating Gradients Through Stochastic Neurons for Conditional Computation},
  author  = {Bengio, Yoshua and L{\'e}onard, Nicholas and Courville, Aaron},
  journal = {arXiv preprint arXiv:1308.3432},
  year    = {2013}
}
}
\clearpage
\appendix
\setcounter{table}{0}
\setcounter{figure}{0}
\setcounter{equation}{0}
\renewcommand{\thetable}{S\arabic{table}}
\renewcommand{\thefigure}{S\arabic{figure}}
\renewcommand{\theequation}{S\arabic{equation}}
\renewcommand{\theHtable}{supp.\arabic{table}}
\renewcommand{\theHfigure}{supp.\arabic{figure}}
\renewcommand{\theHequation}{supp.\arabic{equation}}
\makeatletter
\setlength{\@fptop}{0pt}
\setlength{\@dblfptop}{0pt}
\makeatother
\maketitlesupplementary

\section{Collected-Coordinate Diagnostic, Extended}
\label{supp:diag}
Table~\ref{tab:supp_diag} extends Table~\ref{tab:diag} of the main paper to the fc1 input and to Swin and Hiera, and adds a third statistic, the ratio. Table~\ref{tab:supp_act} extends the fc2 diagnostic to vision and language models with GELU, ReLU and SwiGLU activations. These tables support two observations in Sec.~\ref{sec:where} of the main paper. In Table~\ref{tab:supp_diag}, at full width the collected coordinate sets the fc2 range in every block, but stays far below the maximum at qkv and fc1. Table~\ref{tab:supp_act} shows the same fc2 pattern at GELU- and ReLU-fed inputs. At the SwiGLU-fed inputs of EVA-02 and Qwen2.5, the channel means cancel and no collected-coordinate outlier appears.

The ratio is the largest magnitude of the collected coordinate over all tokens, divided by the 99.9th percentile of the magnitudes of all other rotated coordinates. It measures how far the collected coordinate lies above the bulk of the input. When the collected coordinate sets a max-scaled range, the other coordinates fill only $1/\text{ratio}$ of that range, and $\log_2(\text{ratio})$ of the 4 bits are spent on the collected coordinate alone. A ratio well above 1 marks the collected coordinate as an outlier. For example, the full-width Hadamard at fc2 reaches 3.07 to 9.08 (Table~\ref{tab:supp_diag}). An ordinary coordinate gives a ratio near 1, as the random rotation does with 1.09 to 1.64, and a ratio below 1 places the collected coordinate inside the bulk.

\paragraph{Definitions.}
Let $\mat{X}\in\mathbb{R}^{M\times K}$ be a layer input of the floating-point model and $\mat{Y}=\mat{X}\mat{H}$ its rotation. For a full-width Hadamard, whose first row is all ones, the collected coordinate is coordinate 0 of $\mat{Y}$, each token's channel sum divided by $\sqrt{K}$. For GH every group of $n$ channels has its own coordinate 0, and the statistics use the largest of these $K/n$ coordinates. For a random orthogonal rotation it is the coordinate most aligned with the direction of the token mean. Over all tokens of a block, share $=\max|\text{collected}|/\max|\mat{Y}|$, so a share of 1.00 means that the collected coordinate sets a max-scaled per-tensor range. Ratio $=\max|\text{collected}|/p_{99.9}|\text{other coordinates}|$, as above. Bits $=\log_2(\max|\mat{Y}|/p_{99.9}|\mat{Y}|)$ over the whole rotated input, the levels of a max-scaled 4-bit grid spent above the 99.9th percentile. The three statistics are computed per block on 8 images with every token and reported as the median over blocks. GH uses 64 channels at fc2 and, at qkv and fc1, the largest power of two dividing $K$, at most 128 (16 to 128 on Hiera). Full-width matrices for widths that are not powers of two are Kronecker products of a small Hadamard matrix and a Sylvester matrix, as in QuIP\# and QuaRot~\cite{quipsharp,quarot}, with signs chosen so that row 0 is all ones.
\begin{table*}[t]
\centering\footnotesize\setlength{\tabcolsep}{3.4pt}
\caption{\textbf{The collected coordinate on ten models} (floating point, median over blocks, 8 images). Columns: input width $K$ (range over stages for Swin and Hiera), bits of the unrotated input, and ratio, share and bits of GH, the full-width Hadamard and a random orthogonal rotation. Share 1.00: the collected coordinate sets a max-scaled range. Plain ViTs on ImageNet images with RepQ-ViT's checkpoints (ViT-B AugReg); Swin with the 22k$\to$1k checkpoints of Table~\ref{tab:plugin} of the main paper; Hiera: the SAM2.1 image encoders on COCO images at 1024 pixels.}
\label{tab:supp_diag}
\begin{tabular}{llr r rrr rrr rrr}
\toprule
& & & none & \multicolumn{3}{c}{grouped H} & \multicolumn{3}{c}{full-width H} & \multicolumn{3}{c}{random}\\
\cmidrule(lr){4-4}\cmidrule(lr){5-7}\cmidrule(lr){8-10}\cmidrule(lr){11-13}
Model & input & $K$ & bits & ratio & share & bits & ratio & share & bits & ratio & share & bits\\
\midrule
DeiT-S & qkv & 384 & 1.25 & 0.85 & 0.60 & 0.55 & 0.26 & 0.18 & 0.58 & 1.15 & 0.76 & 0.52\\
       & fc1 & 384 & 1.62 & 0.89 & 0.64 & 0.53 & 0.31 & 0.20 & 0.56 & 1.26 & 0.81 & 0.62\\
       & fc2 & 1536 & 1.80 & 1.60 & 0.84 & 0.92 & 4.41 & \textbf{1.00} & \textbf{2.01} & 1.18 & 0.64 & 0.89\\
ViT-S  & qkv & 384 & 1.01 & 0.94 & 0.65 & 0.55 & 0.30 & 0.20 & 0.53 & 1.09 & 0.74 & 0.59\\
       & fc1 & 384 & 1.55 & 0.97 & 0.68 & 0.57 & 0.29 & 0.20 & 0.56 & 1.17 & 0.85 & 0.58\\
       & fc2 & 1536 & 1.75 & 1.69 & 0.96 & 0.83 & 3.79 & \textbf{1.00} & \textbf{1.80} & 1.23 & 0.70 & 0.85\\
ViT-B  & qkv & 768 & 1.32 & 1.18 & 0.79 & 0.58 & 0.39 & 0.26 & 0.62 & 1.14 & 0.74 & 0.64\\
       & fc1 & 768 & 1.91 & 1.22 & 0.74 & 0.62 & 0.47 & 0.32 & 0.60 & 1.24 & 0.75 & 0.65\\
       & fc2 & 3072 & 2.22 & 1.64 & 0.91 & 0.89 & 4.79 & \textbf{1.00} & \textbf{2.21} & 1.29 & 0.64 & 0.91\\
DeiT-T & qkv & 192 & 0.64 & 0.86 & 0.60 & 0.53 & 0.18 & 0.12 & 0.54 & 1.11 & 0.74 & 0.53\\
       & fc1 & 192 & 0.67 & 0.86 & 0.59 & 0.52 & 0.22 & 0.16 & 0.49 & 1.20 & 0.86 & 0.50\\
       & fc2 & 768 & 1.30 & 1.56 & 0.83 & 0.83 & 3.13 & \textbf{1.00} & \textbf{1.07} & 1.23 & 0.70 & 0.83\\
\midrule
Swin-S & qkv & 96--768 & 1.49 & 1.07 & 0.70 & 0.59 & 0.31 & 0.21 & 0.58 & 1.21 & 0.79 & 0.62\\
       & fc1 & 96--768 & 0.89 & 1.00 & 0.64 & 0.63 & 0.22 & 0.15 & 0.60 & 1.24 & 0.81 & 0.62\\
       & fc2 & 384--3072 & 1.78 & 1.64 & 0.87 & 0.86 & 3.07 & \textbf{1.00} & \textbf{1.48} & 1.25 & 0.67 & 0.92\\
Swin-B & qkv & 128--1024 & 1.32 & 1.09 & 0.70 & 0.68 & 0.33 & 0.21 & 0.60 & 1.22 & 0.80 & 0.57\\
       & fc1 & 128--1024 & 1.03 & 1.10 & 0.70 & 0.63 & 0.29 & 0.20 & 0.57 & 1.29 & 0.84 & 0.58\\
       & fc2 & 512--4096 & 1.83 & 1.71 & 0.92 & 0.84 & 4.36 & \textbf{1.00} & \textbf{1.98} & 1.36 & 0.72 & 0.90\\
\midrule
Hiera-T & qkv & 96--768 & 0.90 & 1.13 & 0.66 & 0.70 & 0.32 & 0.20 & 0.74 & 1.39 & 0.86 & 0.74\\
        & fc1 & 96--768 & 0.79 & 1.15 & 0.70 & 0.70 & 0.47 & 0.28 & 0.67 & 1.43 & 0.87 & 0.70\\
        & fc2 & 384--3072 & 1.72 & 1.80 & 0.87 & 0.70 & 6.17 & \textbf{1.00} & \textbf{2.48} & 1.56 & 0.81 & 0.96\\
Hiera-S & qkv & 96--768 & 1.00 & 1.08 & 0.64 & 0.74 & 0.35 & 0.21 & 0.76 & 1.39 & 0.82 & 0.72\\
        & fc1 & 96--768 & 0.77 & 1.13 & 0.68 & 0.67 & 0.43 & 0.27 & 0.69 & 1.42 & 0.88 & 0.69\\
        & fc2 & 384--3072 & 1.63 & 1.76 & 0.89 & 0.83 & 5.93 & \textbf{1.00} & \textbf{2.46} & 1.59 & 0.85 & 0.91\\
Hiera-B+ & qkv & 112--896 & 1.04 & 1.30 & 0.79 & 0.72 & 0.39 & 0.23 & 0.75 & 1.44 & 0.79 & 0.78\\
         & fc1 & 112--896 & 0.96 & 1.23 & 0.76 & 0.67 & 0.44 & 0.28 & 0.69 & 1.48 & 0.90 & 0.71\\
         & fc2 & 448--3584 & 2.02 & 2.21 & 0.96 & 0.76 & 8.92 & \textbf{1.00} & \textbf{2.93} & 1.54 & 0.77 & 0.98\\
Hiera-L & qkv & 144--1152 & 0.88 & 1.33 & 0.78 & 0.76 & 0.46 & 0.27 & 0.73 & 1.37 & 0.81 & 0.75\\
        & fc1 & 144--1152 & 1.07 & 1.42 & 0.84 & 0.73 & 0.57 & 0.33 & 0.75 & 1.48 & 0.91 & 0.73\\
        & fc2 & 576--4608 & 2.23 & 2.26 & 0.90 & 0.95 & 9.08 & \textbf{1.00} & \textbf{3.10} & 1.64 & 0.78 & 1.05\\
\bottomrule
\end{tabular}
\end{table*}

\paragraph{Ten models.}
Table~\ref{tab:supp_diag} adds the fc1 input and six encoders to Table~\ref{tab:diag} of the main paper. At the LN-fed inputs qkv and fc1 the three rotations spend the same 0.49 to 0.78 bits on every model. At the fc2 input the full-width Hadamard sets the maximum in every block of every model, with a share of 1.00 in the median block and at least 0.99 in every block. Under the full-width Hadamard, the collected coordinate consumes 1.07 to 2.21 bits of quantization range on the plain ViTs, 1.48 and 1.98 bits on Swin-S and Swin-B, and 2.46 to 3.10 bits on the four Hiera encoders. By comparison, the corresponding quantization-range costs are 0.70 to 0.95 bits for GH and 0.83 to 1.05 bits for the random rotation. With the rotation at qkv and fc1 only, the full-width Hadamard and GH differ by at most 0.9 percentage points in top-1 accuracy on DeiT-S, ViT-S and EVA-02-S, and by 0.48 mIoU points on Hiera-T. With the rotation at fc2 as well, the full-width Hadamard trails GH by 4.5 and 12.1 percentage points in top-1 accuracy on DeiT-S and ViT-S, respectively, and by 8.3 mIoU points on Hiera-T. It ties GH on EVA-02-S and EVA-02-B, whose fc2 inputs have no collected coordinate (Table~\ref{tab:ladder} of the main paper and Table~\ref{tab:supp_eva}).

\begin{table*}[t]
\centering\footnotesize\setlength{\tabcolsep}{3.6pt}
\caption{\textbf{The coordinate follows the activation function} (fc2 or down-projection input, floating point, median over blocks; 8 images, or 8 sequences of 512 WikiText-2 tokens for the language models). $\bar\mu/\sigma$: mean over channels of the channel mean, divided by the median channel standard deviation. Neg.: fraction of channels with a negative mean. Other columns as in Table~\ref{tab:supp_diag}. EVA-02-B has a LayerNorm between the SwiGLU product and fc2; both tensors are listed. CLIP: the OpenAI ViT-B/16 image tower.}
\label{tab:supp_act}
\begin{tabular}{llr rr r rr rrr rr}
\toprule
& & & & & none & \multicolumn{2}{c}{grouped H} & \multicolumn{3}{c}{full-width H} & \multicolumn{2}{c}{random}\\
\cmidrule(lr){6-6}\cmidrule(lr){7-8}\cmidrule(lr){9-11}\cmidrule(lr){12-13}
Model & activation & $K$ & $\bar\mu/\sigma$ & neg. & bits & share & bits & ratio & share & bits & share & bits\\
\midrule
DeiT-S & GELU & 1536 & $-0.225$ & 0.88 & 1.80 & 0.84 & 0.92 & 4.41 & 1.00 & 2.01 & 0.64 & 0.89\\
CLIP ViT-B/16 & QuickGELU & 3072 & $-0.432$ & 0.90 & 2.20 & 0.87 & 0.73 & 8.48 & 1.00 & 3.02 & 0.69 & 0.89\\
GPT-2 & GELU & 3072 & $-0.320$ & 0.88 & 2.40 & 0.80 & 1.13 & 8.09 & 1.00 & 2.96 & 0.75 & 0.93\\
Pythia-410M & GELU & 4096 & $-0.347$ & 0.90 & 3.11 & 0.81 & 1.02 & 9.91 & 1.00 & 3.27 & 0.67 & 1.07\\
OPT-125M & ReLU & 3072 & $+0.381$ & 0.01 & 2.48 & 1.00 & 1.65 & 6.10 & 1.00 & 2.56 & 0.61 & 1.58\\
OPT-1.3B & ReLU & 8192 & $+0.270$ & 0.01 & 2.63 & 1.00 & 1.28 & 8.26 & 1.00 & 3.03 & 0.69 & 1.06\\
\midrule
EVA-02-S 336 & SwiGLU & 1024 & $-0.007$ & 0.51 & 2.55 & 0.79 & 0.97 & 1.00 & 0.60 & 0.82 & 0.50 & 1.26\\
EVA-02-B 448, product & SwiGLU & 2048 & $+0.074$ & 0.51 & 3.23 & 0.93 & 0.43 & 0.96 & 0.65 & 0.57 & 0.94 & 0.72\\
EVA-02-B 448, fc2 input & SwiGLU, LN & 2048 & $+0.001$ & 0.51 & 3.30 & 0.86 & 1.09 & 0.58 & 0.32 & 0.87 & 0.54 & 0.98\\
Qwen2.5-1.5B & SwiGLU & 8960 & $+0.000$ & 0.50 & 3.85 & 0.93 & 1.85 & 1.16 & 0.58 & 1.04 & 0.56 & 1.13\\
\bottomrule
\end{tabular}
\end{table*}

\paragraph{GELU and ReLU against SwiGLU.}
Table~\ref{tab:supp_act} compares the fc2 or down-projection input of GELU and ReLU models with that of SwiGLU models. At every GELU- or ReLU-fed input the share is 1.00 in every block. The full-width Hadamard spends 2.0 to 3.3 bits there, against 0.8 to 1.7 for GH and 0.9 to 1.6 for the random rotation. At the fc2 or down-projection inputs of EVA-02-S, EVA-02-B and Qwen2.5 the share is 0.32 to 0.60. The full-width Hadamard spends 0.82 to 1.04 bits there, no more than GH with 0.97 to 1.85 or the random rotation with 0.98 to 1.26. On Qwen2.5 GH spends more bits than the full-width or the random rotation. Its fc2 or down-projection input has outlier channels on every token, 3.85 bits before rotation, and 64 coordinates do not dilute them.

\begin{table}[t]
\footnotesize
\setlength{\tabcolsep}{5pt}
\caption{\textbf{EVA-02, a SwiGLU model, under RepQ-ViT with only qkv, fc1 and fc2 quantized}, ImageNet top-1 (\%), one calibration draw, 50,000 validation images.}
\label{tab:supp_eva}
\begin{tabular*}{\columnwidth}{@{}l@{\extracolsep{\fill}}cc@{}}
\toprule
Transform & EVA-02-S 336 & EVA-02-B 448\\
\midrule
floating point & 85.72 & 88.26\\
none & 71.08 & 76.22\\
full-width Hadamard & 82.65 & 87.58\\
grouped Hadamard (GH) & 82.66 & 87.28\\
random orthogonal & 83.15 & 87.61\\
S, GH & 83.28 & 87.54\\
S, full-width Hadamard & 83.69 & 87.50\\
S, GH with 128-wide fc2 groups & 83.42 & 87.65\\
\bottomrule
\end{tabular*}
\end{table}

\section{Group Width Ablation}
\label{supp:ladder}
Table~\ref{tab:supp_width} reports the measurements underlying the accuracy plateau from 64 to 384 channels discussed in Sec.~\ref{sec:setup} of the main paper. All rows use the W4A4 protocol of Table~\ref{tab:ladder} of the main paper. The host is RepQ-ViT, the calibration draw is draw 0 with 32 ImageNet training images, the evaluation uses the full 50,000 validation images, and each table comes from one machine. Paired differences compare the per-image predictions of two rows over the 50,000 images with the McNemar statistic and carry 95\% intervals.

\begin{table*}[t]
\centering\footnotesize\setlength{\tabcolsep}{5pt}
\caption{\textbf{Group width}, ImageNet top-1 (\%) with GH only and RepQ-ViT's quantizers. Top: fc2 width with qkv and fc1 at 128; 1536 is the full width. Bottom: qkv and fc1 width with fc2 at 64.}
\label{tab:supp_width}
\begin{tabular}{lccccccccc}
\toprule
fc2 width & 16 & 32 & 64 & 128 & 256 & 384 & 512 & 768 & 1536\\
\midrule
DeiT-S & 69.09 & 73.49 & 74.57 & 74.48 & 74.61 & 74.66 & 74.22 & 73.14 & 70.59\\
ViT-S & 72.39 & 73.00 & 72.61 & 72.39 & 72.91 & 72.39 & 71.73 & 68.10 & 60.92\\
\midrule
qkv and fc1 width & & 32 & 64 & 128 & & & & & full\\
\midrule
DeiT-S & & 73.13 & 74.12 & 74.57 & & & & & 74.69\\
ViT-S & & 72.99 & 72.81 & 72.61 & & & & & 72.95\\
\bottomrule
\end{tabular}
\end{table*}

At fc2, ImageNet top-1 accuracy is nearly flat for group widths of 64 to 384 channels: 74.48\% to 74.66\% on DeiT-S and 72.39\% to 72.91\% on ViT-S (Table~\ref{tab:supp_width}). Wider groups reduce accuracy; full-width groups (1536 channels) lose 4.0 and 11.7 percentage points relative to 64-channel groups on DeiT-S and ViT-S, respectively. Groups of 16 and 32 channels also perform worse on DeiT-S. We choose 64 channels to limit fusion cost: it is the narrowest width on the plateau and the largest power of two dividing every model's fc2 width. At qkv and fc1, 128-channel groups keep top-1 accuracy within 0.4 percentage points of the full-width rotation on both models.

\paragraph{Random signs on GH.}
Under RepQ-ViT without S or A, adding random signs to GH changes ImageNet top-1 accuracy by $+0.25$ and $+0.36$ percentage points on DeiT-S, and $+0.33$ and $-0.21$ on ViT-S, across two sign seeds relative to GH without random signs. These changes are small and not consistently positive, so we use GH without random signs by default.

\section{SmoothQuant Scaling Against the Grouped Rotation}
\label{supp:sq}
This section compares SmoothQuant's per-channel scaling with GH under the same W4A4 plug-in protocol on RepQ-ViT (Table~\ref{tab:supp_sq}).

\begin{table*}[t]
\centering\footnotesize\setlength{\tabcolsep}{6pt}
\caption{\textbf{SmoothQuant scaling versus GH on RepQ-ViT.} ImageNet top-1 accuracy (\%) under the W4A4 plug-in protocol of Table~\ref{tab:plugin} of the main paper. none (RepQ-ViT): the original baseline with LayerNorm reparameterization; none (per-tensor): the same per-tensor input quantizer configuration as SmoothQuant and GH at qkv, fc1 and fc2, with neither scaling nor rotation; SmoothQuant: per-channel scaling with $\alpha=0.5$ at these three inputs. Swin-B uses the zero-point fix.}
\label{tab:supp_sq}
\begin{tabular}{lrrrrrrr}
\toprule
Transform & ViT-S & ViT-B & DeiT-T & DeiT-S & DeiT-B & Swin-S & Swin-B\\
\midrule
none (RepQ-ViT) & 65.24 & 72.13 & 57.95 & 68.93 & 75.68 & 79.69 & 80.71\\
none (per-tensor) & 61.71 & 72.22 & 50.23 & 44.06 & 73.40 & 79.68 & 80.18\\
SmoothQuant & 60.23 & 60.68 & 50.53 & 51.82 & 75.01 & 76.28 & 74.80\\
GH & 72.61 & 81.08 & 62.40 & 74.57 & 79.37 & 81.23 & 82.45\\
\bottomrule
\end{tabular}
\end{table*}

For the SmoothQuant baseline, we replace the rotation with per-channel scaling~\cite{smoothquant}:
\[
s_c = \frac{\bigl(\max_t |x_{tc}|\bigr)^{\alpha}}{\bigl(\max_o |W_{oc}|\bigr)^{1-\alpha}}.
\]
Here, $s_c$ is the scale for input channel $c$, $x_{tc}$ is the floating-point input activation of token $t$ in channel $c$, and $W_{oc}$ is the pretrained weight connecting input channel $c$ to output channel $o$. The activation maximum is taken over all tokens in the 32 calibration images, and the weight maximum over all output channels. The exponent $\alpha\in[0,1]$ controls the balance between activation and weight magnitudes in the scale; we set $\alpha=0.5$ in these comparisons. At qkv, fc1 and fc2, we divide input channel $c$ by $s_c$ before per-tensor activation quantization and multiply the corresponding weight column by $s_c$. SmoothQuant, GH and the no-transform per-tensor baseline use the same quantizer configurations. The original RepQ-ViT baseline is reported separately because it retains LayerNorm reparameterization.

At $\alpha=0.5$, SmoothQuant trails GH by 4.4 to 22.8 percentage points in ImageNet top-1 accuracy across all seven models; all paired 95\% confidence intervals for these differences exclude zero. Compared with the no-transform per-tensor baseline, SmoothQuant improves top-1 accuracy by 7.8 percentage points on DeiT-S and 1.6 percentage points on DeiT-B. The 0.3-percentage-point increase on DeiT-T is not statistically significant ($p=0.15$). On ViT-S, ViT-B, Swin-S and Swin-B, it instead reduces top-1 accuracy by 1.5 to 11.5 percentage points. In separate W4A4 simulations on the SAM2.1 Hiera-T and Hiera-L encoders, SmoothQuant with $\alpha=0.5$ trails GH by 0.9 and 7.5 mIoU points, respectively.

\section{Plug-in Details}
\label{supp:imagenet}
We give here the implementation details of S and A on RepQ-ViT and the experimental controls for the host methods in Table~\ref{tab:plugin} of the main paper. For the ImageNet hosts, we use the released calibration and reconstruction configurations, with the GH integration described in Sec.~\ref{sec:plugin} of the main paper. For these ImageNet hosts, we follow the evaluation protocol and image preprocessing of their released implementations and report top-1 accuracy on all 50,000 ImageNet validation images.

\paragraph{S and A on RepQ-ViT.}
On RepQ-ViT, we use 512 additional ImageNet training images, separate from the 32 images used for host quantizer calibration, to estimate the S statistics and fit A. S uses floating-point input statistics from these images. A uses cached floating-point block inputs and outputs and the relative output MSE of Eq.~(\ref{eq:loss}) of the main paper. The fit runs 1,000 steps with one image per step, AdamW at a learning rate of $5\times10^{-3}$, zero weight decay, cosine decay and gradient-norm clipping at 1. Weights and calibrated host quantizers stay fixed. During A fitting, we keep the host quantizers' forward computations unchanged and use a straight-through estimator~\cite{bengio2013ste} to pass gradients through rounding. Gradients pass through inputs within the quantization range and are blocked for inputs clipped to its boundaries. On this host A corrects all quantized linear layers within the blocks, including proj and the Swin patch-merging reduction, which receive A alone with their host input quantization retained. The patch embedding, the classification head and the attention products receive no transform. On our machine with a single RTX 4090, fitting A takes 95 s for ViT-B and 202 s for Swin-S.

\paragraph{Checkpoints.}
To ensure a fair comparison, we use the same pretrained checkpoint for each ImageNet model across all four reproduced host methods and their plug-in variants in Table~\ref{tab:plugin} of the main paper. Specifically, we use ViT-B AugReg2 and the ImageNet-22k-pretrained, ImageNet-1k-finetuned checkpoints for Swin-S and Swin-B, following the checkpoint choices of FIMA-Q and LS-ViT. Results quoted from prior papers may use different checkpoints; Table~\ref{tab:supp_ckpt} summarizes these differences for ViT-B and Swin-S.

\begin{table}[t]
\centering\footnotesize\setlength{\tabcolsep}{3pt}
\caption{\textbf{ViT-B and Swin-S checkpoints of the published ImageNet results} in Table~\ref{tab:plugin} of the main paper, and our reproduction of each method on that checkpoint (top-1, ours / published).}
\label{tab:supp_ckpt}
\begin{tabular}{lllcc}
\toprule
& \multicolumn{2}{c}{Checkpoint} & \multicolumn{2}{c}{Top-1, ours / published}\\
\cmidrule(lr){2-3}\cmidrule(lr){4-5}
Method & ViT-B & Swin-S & ViT-B & Swin-S\\
\midrule
RepQ-ViT & AugReg & 1k & 68.24 / 68.48 & 79.54 / 79.45\\
I\&S-ViT & AugReg & 1k & 80.16 / 80.07 & 81.11 / 81.17\\
FIMA-Q & AugReg2 & 22k$\to$1k & 82.96 / 83.04 & 81.76 / 81.82\\
LS-ViT & AugReg2 & 22k$\to$1k & 83.05 / 83.08 & 81.81 / 81.82\\
Others$^\ast$ & \multicolumn{4}{l}{not verified}\\
\bottomrule
\multicolumn{5}{l}{\scriptsize $^\ast$PTQ4ViT, ERQ, APHQ-ViT and RepQ-ViT + QwT.}\\
\end{tabular}
\end{table}

\paragraph{Zero-point defect of RepQ-ViT on Swin-B.}
In our Swin-B reproduction, a nearly inactive LayerNorm channel produces an extremely small quantization step, causing an out-of-range zero point in RepQ-ViT's reparameterization. We fix this by setting a lower bound of 1\% of the median channel step before reparameterization. The fix raises ImageNet top-1 accuracy from 70.48\% to 80.71\%, while leaving the GH variants unchanged (Table~\ref{tab:supp_repqfix}). Table~\ref{tab:plugin} of the main paper uses the corrected Swin-B baseline and the original RepQ-ViT implementation for the other six models.

\begin{table}[t]
\centering\footnotesize\setlength{\tabcolsep}{3pt}
\caption{\textbf{RepQ-ViT W4A4 top-1 with the released code and with the zero-point fix} (released / fixed), on the checkpoints and calibration draw of Table~\ref{tab:plugin} of the main paper. Our rows are identical with and without the fix. Gains are against the fixed base.}
\label{tab:supp_repqfix}
\begin{tabular}{lcccc}
\toprule
Model & RepQ-ViT & + GH & + GH, S, A & Gains\\
\midrule
ViT-S & 65.24 / 64.56 & 72.61 & 75.73 & +8.05, +11.17\\
ViT-B & 72.13 / 72.26 & 81.08 & 82.33 & +8.82, +10.07\\
DeiT-T & 57.95 / 57.78 & 62.40 & 65.03 & +4.61, +7.25\\
DeiT-S & 68.93 / 68.87 & 74.57 & 76.05 & +5.70, +7.18\\
DeiT-B & 75.68 / 75.50 & 79.37 & 80.19 & +3.87, +4.69\\
Swin-S & 79.69 / 79.71 & 81.23 & 81.48 & +1.51, +1.77\\
Swin-B & 70.48 / 80.71 & 82.45 & 83.19 & +1.74, +2.48\\
\bottomrule
\end{tabular}
\end{table}

\paragraph{Calibration draws.}
Table~\ref{tab:supp_multidraw} evaluates the stability of GH gains across three calibration draws for the eight host--model pairs with the smallest GH gains over their respective baselines in Table~\ref{tab:plugin} of the main paper. The reproduced ImageNet results in the main table use one draw per pair. Within each draw, the baseline and GH use the same calibration images and cached crops, with matched random seeds for reconstruction.

GH improves ImageNet top-1 accuracy in every draw for all eight pairs. Its mean gains range from 0.18 to 0.60 percentage points, with sample standard deviations no greater than 0.10 percentage points.

To account for the same validation images being used across draws, we first average each image's paired correctness differences (GH minus baseline) over the three fitted draws. We then resample the 50,000 images with replacement 100,000 times and use the 2.5th and 97.5th percentiles of the resampled mean differences as the 95\% confidence interval. These intervals reflect image-sampling uncertainty conditional on the three fitted draws. All eight intervals lie above zero; the smallest and largest mean gains are $0.18$ (95\% CI: $[0.07,0.30]$) and $0.60$ ($[0.49,0.72]$) percentage points, respectively. A two-sided normal-approximation test, using the mean paired difference and its standard error across images, gives $p<0.002$ for every pair.

Of the 24 individual-draw comparisons, 23 are significant at $p<0.05$. The only exception is the first draw of I\&S-ViT on Swin-S, where GH gains 0.10 percentage points ($p=0.32$).

In separate RepQ-ViT checks, redrawing both the calibration and compensation images changes the top-1 accuracy of GH with S and A by $+0.15$, $+0.11$, $-0.10$ and $+0.04$ percentage points on ViT-B (AugReg), DeiT-T, DeiT-S and Swin-T, respectively. On Swin-B, GH with S and A achieves 83.19\%, 83.24\% and 83.37\% top-1 accuracy across three draws, whereas the uncorrected RepQ-ViT baseline achieves 70.48\%, 69.26\% and 70.49\%, respectively.

\begin{table}[t]
\centering\footnotesize\setlength{\tabcolsep}{3pt}
\caption{\textbf{GH stability across three calibration draws.} The eight host--model pairs with the smallest GH gains in Table~\ref{tab:plugin} of the main paper. Baseline and + GH: ImageNet top-1 accuracy (\%); GH gain: paired difference in percentage points. Values are mean $\pm$ sample standard deviation across draws.}
\label{tab:supp_multidraw}
\begin{tabular*}{\columnwidth}{@{}l@{\extracolsep{\fill}}lrrr@{}}
\toprule
Host & Model & Baseline & + GH & GH gain\\
\midrule
I\&S-ViT & Swin-S & 81.27$\pm$0.08 & 81.45$\pm$0.05 & +0.18$\pm$0.08\\
I\&S-ViT & Swin-B & 82.55$\pm$0.09 & 83.08$\pm$0.08 & +0.52$\pm$0.04\\
FIMA-Q & DeiT-B & 80.23$\pm$0.03 & 80.69$\pm$0.02 & +0.46$\pm$0.02\\
FIMA-Q & Swin-S & 81.73$\pm$0.06 & 82.12$\pm$0.08 & +0.38$\pm$0.08\\
FIMA-Q & Swin-B & 83.46$\pm$0.04 & 84.06$\pm$0.05 & +0.60$\pm$0.07\\
LS-ViT & DeiT-B & 80.31$\pm$0.10 & 80.81$\pm$0.04 & +0.50$\pm$0.06\\
LS-ViT & Swin-S & 81.76$\pm$0.06 & 82.11$\pm$0.06 & +0.35$\pm$0.10\\
LS-ViT & Swin-B & 83.50$\pm$0.08 & 84.05$\pm$0.01 & +0.55$\pm$0.08\\
\bottomrule
\end{tabular*}
\end{table}

\paragraph{Direct AHCQ-SAM integration.}
The SAM v1 plug-in uses AHCQ-SAM revision \texttt{a43f427}, its W4A4 configuration, its first 32 COCO training images and seed 20261010. Before running AHCQ-SAM, we calibrate and quantize the qkv, fc1 and fc2 layers of every encoder block using activations from the floating-point model. We apply the quantizers of Table~\ref{tab:samv1} of the main paper (Sec.~\ref{supp:samv1}) only to these three inputs and keep these layers frozen during AHCQ-SAM. AHCQ-SAM then calibrates, quantizes and reconstructs everything else unchanged. For the AHCQ-SAM baseline without GH, calibration and reconstruction take 28.4, 39.4 and 58.0 hours for SAM-B, SAM-L and SAM-H, respectively, on a single RTX 4090 with 48 GB of memory, excluding evaluation. %

\paragraph{SAM2.1 under AHCQ-SAM.}
The video plug-in follows the full released scope of AHCQ-SAM on SAM2.1 Tiny. We follow AHCQ-SAM's released calibration and evaluation protocol and report frame-weighted J\&F on the 150 SA-V test videos. After adding GH, J\&F rises from 50.13 to 54.64, J from 47.99 to 51.82, and F from 52.27 to 57.46. A paired video-cluster bootstrap with 10,000 draws gives the gain the interval $[+2.76,+6.28]$, with 114 of the 150 videos up and 34 down.

\section{SAM2.1 Engine Evaluation and Ablations}
\label{supp:sam21}
This section describes the SAM2.1 point-prompt evaluation used for the native INT4 engine and its associated simulations and ablations in Secs.~\ref{sec:sam21} and~\ref{sec:ablation} of the main paper. These experiments measure instance-weighted mIoU on COCO images; the AHCQ-SAM video plug-in uses the separate SA-V protocol described in Sec.~\ref{supp:imagenet}.

\paragraph{Protocol and fitting.}
The point protocol uses the first 1,000 sorted COCO validation image IDs. Instances have an area above 100 pixels, at most eight per image, and 992 images contribute 4,629 instances. Each instance gets one centroid-interior point. The mask is selected by predicted IoU. The prompt encoder and the mask decoder run in floating point, and mIoU is instance-weighted. Native encoder features enter this same pipeline. Fits use a fixed set of 32 COCO training images. The main-table seeds are 20261010 through 20261014, and the ablations use the first two. The native W4A4 path uses symmetric codes in $[-7,7]$, per-output-channel weight scales, per-token activation scales at qkv and fc1 and one static scale per layer at fc2. A and the clipping ratios are fitted jointly against floating-point block outputs for 1,000 AdamW steps with one image per step. The learning rates are $5\times10^{-3}$ for A and $5\times10^{-2}$ for the clipping ratios, with cosine decay. S stays in closed form unless varied explicitly.

\paragraph{Paired mIoU intervals.}
For the SAM2.1 point-prompt comparisons, we compute paired 95\% confidence intervals as follows. Each instance's IoU difference between a row and its same-seed reference is first averaged over the seeds present in both rows. We resample the 992 images with replacement 10,000 times, keeping their instances. Each replicate is instance-weighted, and the 2.5th and 97.5th percentiles give the interval. It measures the image-sampling uncertainty for the fitted seeds.

\section{SAM v1 Encoder Quantization}
\label{supp:samv1}
This section describes the separate SAM v1 encoder experiments in Table~\ref{tab:samv1} of the main paper. They evaluate GH under the full encoder quantization scope of PTQ4SAM using COCO mask AP with YOLOX and DINO box prompts, rather than the point-prompt mIoU protocol of Sec.~\ref{supp:sam21}.

\paragraph{Quantization and fitting.}
Table~\ref{tab:samv1} of the main paper quantizes every encoder operand of SAM ViT-B, ViT-L and ViT-H at W4A4 in the PTQ4SAM code. The decoder runs in floating point, and the first 32 COCO training images serve for calibration. GH acts at every GEMM input, qkv, proj, fc1, fc2 and the neck, and both rows keep the per-head Hadamard on Q, K and V of that code. The weights are rounded to nearest with one scale per output channel and a learned clipping ratio per row. At qkv and fc1 the activation scale is dynamic per token, the token's maximum times a learned ratio. At fc2 it is one static symmetric scale per layer, a learned ratio times the maximum of the rotated input over the 32 calibration images. The ratios are learned per encoder block for 1,000 AdamW steps with the loss of Eq.~(\ref{eq:loss}) of the main paper, with S and A off. The none row switches off GH at the encoder GEMM inputs while retaining the per-head Hadamard on Q, K and V, with the same quantizers and fits. The fits take 6, 14 and 26 minutes on B, L and H on one RTX 4090.

\section{Latency on an RTX 3090}
\label{supp:runtime}

\begin{table}[t]
\centering\footnotesize\setlength{\tabcolsep}{4pt}
\caption{\textbf{Latency on an RTX 3090}, ms per batch, the protocol of Table~\ref{tab:main} of the main paper on a second GPU: the same engine built for sm\_86, and TensorRT 11.2 FP16 plans built on that card. Ratio: W8A8 over W4A4, and TensorRT FP16 over W4A4.}
\label{tab:supp_3090}
\begin{tabular}{l rrr rr}
\toprule
& \multicolumn{3}{c}{ours} & \multicolumn{2}{c}{TensorRT}\\
\cmidrule(lr){2-4}\cmidrule(lr){5-6}
Encoder, batch & W8A8 & W4A4 & ratio & FP16 & ratio\\
\midrule
Hiera-T, 1 & 4.92 & 4.34 & 1.13 & 8.07 & 1.86\\
Hiera-S, 1 & 5.78 & 5.02 & 1.15 & 9.36 & 1.87\\
Hiera-B+, 1 & 8.90 & 7.38 & 1.21 & 14.21 & 1.93\\
Hiera-L, 1 & 20.56 & 15.83 & 1.30 & 31.64 & 2.00\\
Hiera-T, 8 & 32.44 & 28.19 & 1.15 & 62.45 & 2.22\\
Hiera-S, 8 & 38.33 & 32.65 & 1.17 & 71.66 & 2.20\\
Hiera-B+, 8 & 64.32 & 52.50 & 1.23 & 117.65 & 2.24\\
Hiera-L, 8 & 151.88 & 116.86 & 1.30 & 262.80 & 2.25\\
\bottomrule
\end{tabular}
\end{table}

Table~\ref{tab:supp_3090} gives the SAM2.1 encoder latency of the W4A4 engine and its baselines on an RTX 3090, a Linux host at 350 W, under the protocol of Table~\ref{tab:main} of the main paper. Each cell is timed in one session with CUDA graphs, 200 warm-up and 200 measured iterations, as the median of six rounds balanced in configuration order, with the card idle below 50\,$^\circ$C before every cell. The engine is the same source built for sm\_86 and tuned with the rule used for the 4090 build, which leaves the 4090 kernels unchanged. The TensorRT 11.2 FP16 plans are built on that card. The W4A4 encoders run 1.13 to 1.30 times faster than W8A8 and 1.86 to 2.25 times faster than TensorRT FP16. On Hiera-L the ratio over W8A8 is 1.30 at both batch sizes, against 1.27 and 1.26 on the RTX 4090. On Hiera-T and Hiera-S at batch 1 it is 0.04 below the 4090. Accuracy is measured on the RTX 4090 only.

\end{document}